\documentclass{aamas2016}
\usepackage{times}
\usepackage{helvet}
\usepackage{courier}
\usepackage{amsmath,amssymb}
\usepackage{epsfig}
\usepackage{graphicx}
\usepackage{url}
\usepackage[hang,raggedright]{subfigure}
\usepackage{enumitem}
\setenumerate{noitemsep,topsep=3pt,parsep=3pt,partopsep=1pt}
\setitemize{noitemsep,topsep=3pt,parsep=3pt,partopsep=1pt}

\newtheorem{lemma}{Lemma}
\newtheorem{theorem}{Theorem}
\newtheorem{definition}{Definition}

\newcommand{\attackletter}{\mathrm{t}}

\DeclareRobustCommand{\eulerian}{\genfrac\langle\rangle{0pt}{}}

\usepackage[flushmargin]{footmisc}

\begin{document}
\toappear{}
\sloppy
%
\title{Learning Adversary Behavior in Security Games: A PAC Model Perspective}
\author{
\alignauthor
Arunesh Sinha, Debarun Kar, Milind Tambe\\ 
\affaddr{University of Southern California}
\email{\{aruneshs, dkar, tambe\}@usc.edu}
}

\maketitle
\begin{abstract}
Recent applications of Stackelberg Security Games (SSG),
from wildlife crime to urban crime, have employed machine
learning tools to learn and predict adversary behavior
using available data about defender-adversary
interactions. Given these recent developments, this paper commits to an approach of directly
learning the response function of the adversary. Using the PAC model,
this paper lays a firm theoretical foundation for learning in SSGs (e.g., theoretically
answer questions about the numbers of samples required to learn adversary behavior) and 
provides utility guarantees when the learned adversary model is used to plan the defender's strategy.
The paper also aims to answer practical questions such as how much more
data is needed to improve an adversary model's accuracy. Additionally, we explain a recently
observed phenomenon that prediction accuracy
of learned adversary behavior is not enough to discover
the utility maximizing defender strategy. We
provide four main contributions: (1) a PAC model of learning adversary response functions in SSGs;
(2) PAC-model analysis of the learning of key, existing bounded rationality
models in SSGs; (3) an entirely new approach to adversary modeling based on
a non-parametric class of response functions with PAC-model analysis and
(4) identification of conditions under which computing the best
defender strategy against the learned adversary behavior
is indeed the optimal strategy. Finally, we conduct
experiments with real-world data from a national park
in Uganda, showing the benefit of our new adversary modeling approach
and verification of our PAC model predictions.
\end{abstract}
\section{Introduction}
Stackelberg Security Games (SSGs) are arguably the best example of the application of the Stackelberg game model in the real world. Indeed, numerous successful deployed applications~\cite{Tambe2011} (LAX airport, US air marshal) 
and extensive research on related topics~\cite{basilico2009leader,Korzhyk10,gan2015security} provide evidence about the generality of the SSG framework. More recently, new application domains of SSGs, from wildlife crime to urban crime,
are accompanied by significant amounts of past data of recorded defender strategies and adversary reactions.
This has enabled the learning of adversary behavior from such data~\cite{Zhang15a,Yang20142615805}. Also, analysis of these datasets and human subject experiment studies~\cite{Nguyen13analyzingthe} have revealed that modeling bounded rationality of the adversary enables the defender to further optimize her allocation of limited security resources. Thus, learning the adversary's bounded rational behavior and computing defender strategy based on the learned model has become an important area of research in SSGs.

However, without a theoretical foundation for this learning and strategic planning problem, many issues that arise in practice cannot be explained or addressed. For example, it has been recently observed that in spite of good prediction accuracy of the learned models of adversary behavior, the performance of the defender strategy that is computed against this learned adversary model is poor~\cite{Ford15a}. A formal study could also answer several other important questions that arise in practice, for example, (1) How many samples would be required
to learn a ``reasonable'' model of adversary behavior in a given SSG? 
(2) What utility bound can
be provided when deploying the best defender strategy that is computed
against the learned adversary model?

Motivated by the learning of adversary behavior from data in recent applications~\cite{haskell2014robust,Zhang15a}, we adopt the framework in which the defender first learns the \emph{response function} of the adversary (adversary behavior) and then optimizes against the learned response. This paper is the first theoretical study of the adversary bounded rational behavior learning problem and the optimality guarantees (utility bounds) when  computing the best defender strategies against such learned behaviors. 
Indeed, unlike past theoretical work on learning in SSGs (see related work) where reasoning about adversary response happens through payoff and rationality, we treat the response of the bounded rational adversary as the object to be learned. 

Our \emph{first contribution} is using the Probably Approximately Correct (PAC) model~\cite{kearns1994introduction,Anthony2009NNL} to analyze the learning problem at hand. A PAC analysis yields sample complexity, i.e., the number of samples required to achieve a given level of learning guarantee. Hence, the PAC analysis allows us to address the question of required quantity of samples raised earlier. While PAC analysis is fairly standard for classifiers and real valued functions (i.e., regression) it is not an out-of-the-box approach.  In particular, PAC-model analysis of SSGs brings to the table significant new challenges. To begin with, given that we are learning adversary response functions, we 
must deal with the output being a probability
distribution over the adversary's actions, i.e., these response functions are vector-valued. We appeal to the framework of Haussler~\cite{Haussler1992} to study the PAC learnability of \emph{vector-valued} response functions. For SSGs, we first pose the learning problem in terms of maximizing the likelihood of seeing the attack data, but without restricting the formulation to any particular class of response functions. This general PAC framework for learning adversary behavior in SSGs enables the rest of the analysis in this paper.


Our \emph{second contribution} is an analysis of the SUQR model of bounded rationality adversary behavior used in SSGs, which posits a class of parametrized response functions with a given number of parameters (and corresponding features). SUQR is the best known model of bounded rationality in SSGs, resulting in multiple deployed applications~\cite{haskell2014robust,fang2015gsg}. In analyzing SUQR, we advance the state-of-the-art in the mathematical techniques involved in PAC analysis of vector-valued function spaces. In particular, we provide a technique to obtain sharper sample complexity for SUQR than simply directly applying Haussler's original techniques. We decompose the given SUQR function space into two (or more) parts, performing PAC analysis of each part and finally combining the results to obtain the sample complexity result (which scales as $T \log T$ with $T$ targets) for SUQR (see details in Section~\ref{samplecomplexity}). 

Our \emph{third contribution} includes an entirely new behavioral model specified by the \emph{non-parametric Lipschitz} (NPL) class of response functions for SSGs, where the only restriction on NPL functions is Lipschitzness. The NPL approach makes very few assumptions about the response function, enabling the learning of a multitude of behaviors albeit at the cost of higher sample complexity. As NPL has never been explored in learning bounded rationality models in SSGs, we provide a \emph{novel learning technique} for NPL. We also compute the sample complexity for NPL.
Further, we observe in our experiments that the power to capture a large variety of behaviors enables NPL to perform better than SUQR with the real-world data from Queen Elizabeth National Park (QENP) in Uganda. 

Our \emph{fourth contribution} is to convert the PAC learning guarantee into a bound on the utility derived by the defender when planning her strategy based on the learned adversary behavior model. In the process, we make explicit the assumptions required from the dataset of adversary's attacks in response to deployed defender mixed strategies in order to discover the optimal (w.r.t. utility) defender strategy. 
These assumptions help explain a puzzling phenomenon observed in recent literature on learning in SSGs~\cite{Ford15a}---in particular that learned
adversary behaviors provide good prediction accuracy, but the best defender
strategy computed against such learned behavior may not perform well in practice.
The key is that the dataset for learning must not simply record a large number of attacks against few defender strategies
but rather contain the attacker's response against a variety of defender
mixed strategies.  We discuss the details of our assumptions and its implications for the strategic choice of defender's actions in Section~\ref{utilitybounds}.

We also conduct experiments with real-world poaching data from the QENP in Uganda (obtained from \cite{Nguyen15a}) and data collected from human subject experiments. The experimental results support our theoretical conclusions about the number of samples required for different learning techniques. Showing the value of our new NPL approach, the NPL approach outperforms all existing approaches in predicting the poaching activity in QENP.  Finally, our work opens up a number of exciting research directions, such as  studying learning of behavioral models in active learning setting and real-world application of non-parametric models. 
\footnote{Due to lack of space, some proofs in this paper are in the online Appendix: \url{http://bit.ly/1l4n3s1}}

\section{Related Work}
Learning and planning in SSGs with rational adversaries has been studied in two recent papers~\cite{blum2014learning,Balcan2015}, and in Stackelberg games by Letchford et al.~\shortcite{letchford2009learning} and Marecki et al.~\shortcite{marecki2012playing}. All these papers study the learning problem under an active learning framework, where the defender can choose the strategy to deploy within the learning process. Also, all these papers study the setting with perfectly rational adversaries. Our work differs as we study bounded rational adversaries in a passive learning scenario (i.e., with given data) and once the model is learned we analyze the guarantees of planning against the learned model. 
Also, our focus on SSGs differentiates us from recent work on PAC learnability in co-operative games~\cite{balcan2015learning}, in which the authors study PAC learnability of the value function of coalitions with perfectly rational players.
Also, our work is orthogonal to adversarial learning~\cite{vorobeychik2014optimal}, which studies game theoretic models of an adversary attacking a learning algorithm.

PAC learning model has a very rich and extensive body of work~\cite{Anthony2009NNL}. The PAC model provides a theoretical underpinning for most standard machine learning techniques. We use the PAC framework of Haussler~\cite{Haussler1992}. For the parametric case, we derive sharp sample complexity bounds based on covering numbers using our techniques rather than bounding it using the standard technique of pseudo-dimension~\cite{pollard1984convergence} or fat shattering dimension~\cite{Anthony2009NNL}. For the NPL case we use results from~\cite{tikhomirov1993varepsilon} along with our technique of bounding the mixed strategy space of the defender; these results \cite{tikhomirov1993varepsilon} have also been used in the study of Lipschitz classifiers~\cite{luxburg2004distance} but we differ as our hypothesis functions are real vector-valued. 

\section{SSG Preliminaries}
This section introduces the background and preliminary notations for SSGs. A summary of notations used in this paper is presented in Table \ref{Table1}. 
An SSG 
is a two player Stackelberg game between a defender (leader) and an adversary (follower)~\cite{Paruchuri2008}. 
The defender wishes to protect $T$ targets with a limited number of security resources $K$ ($K <\!< T$). 
For ease of presentation, we restrict ourselves to the scenario with no scheduling constraints (see Korzhyk et al.~\shortcite{Korzhyk10}). The defender's pure strategy is to allocate each resource to a target. 
A defender's mixed-strategy $\tilde x$ ($\forall j \in \mathcal{P}.~  \tilde x_{j}\in [0,1], \sum_{j=1}^{\mathcal{P}}\tilde x_{j}=1)$ is then defined as a probability distribution over the set of all possible pure strategies $\mathcal{P}$. An equivalent description (see Korzhyk et al.~\shortcite{Korzhyk10}) of these mixed strategies are coverage probabilities over the set of targets: $x$ ($ \forall i \in T.~x_{i}\in [0,1], \sum_{i=1}^{T}x_{i}\leq K)$. We refer to this latter description as the mixed strategy of the defender.

A pure strategy of the adversary is defined as attacking a single target. The adversary's mixed strategy is then a categorical distribution over the set of targets. Thus, it can be expressed as parameters $q_i$ $(i\in T)$ of a categorical distribution such that $0 \leq q_i \leq 1$ and $\sum_i q_i = 1$.  The adversary's response to the defender's 
mixed strategy is given by a function $q: X \rightarrow Q$, where $Q$ is the space of all mixed strategies of the adversary. The matrix $U$ specifies the payoffs of the defender, and her expected utility is $x^T U q(x)$ when she plays a 
mixed strategy $x \in X$. 

\textbf{Bounded Rationality Models:}
We discuss the SUQR model and its representation for the analysis in this paper below.
Building on prior work on quantal response~\cite{McFadden1976}, SUQR~\cite{Nguyen13analyzingthe} states that given $n$ actions, a human player plays action $i$ with probability $q_i \propto e^{w \cdot v}$, where $v$ denotes a vector of feature values for choice $i$ and $w$ denotes the weight parameters for these features. 
The model is equivalent to conditional logistic regression~\cite{mcfadden1973conditional}. The features are specific to the domain, e.g., in case of SSG applications, the set of features include the coverage probability $x_i$, the reward $R_i$ and penalty $P_i$ of target $i$. Since, other than the coverage $x$, remaining features are fixed for each target in real world data, we assume a target-specific feature $c_i$ (which may be a linear combination of rewards and penalties) and analyze the following \emph{generalized}\footnote{This general form is harder to analyze than the \emph{standard} SUQR form in which the exponent function (function of $x_i, R_i, P_i$) for all $q_i$ is same: $w_1 x_i + w_2 R_i + w_3 P_i$. For completeness, we derive the results for the standard SUQR form in the Appendix.} form of SUQR with parameters $w_1$ and $c_i$'s: $q_i (x) \propto e^{w_1 x_i + c_i}$. As $\sum_{i=1}^T q_i(x) = 1$, we have: 
\vspace{-5pt}
$$q_i(x) = \frac{e^{w_1 x_i + c_i}}{ \sum_{j=1}^{T} e^{w_1 x_j + c_j}}.$$

\noindent \textbf{Equivalent Alternate Representation}:
For ease of mathematical proofs, using standard techniques in logistic regression, we take $q_T \propto e^0$, and hence, $q_i \propto e^{w_1 (x_i - x_T) + (c_i - c_T)}$. To shorten notation, let $c_{iT} = c_i - c_T$, $x_{iT} = x_i - x_T$.  By multiplying the numerator and denominator by $e^{w_1 x_T +  c_T}$, it can be verified that $q_i(x) = \frac{e^{w_1 x_{iT} + c_{iT}}}{ e^0 + \sum_{j=1}^{T-1} e^{w_1 x_{jT} + c_{jT}}} = \frac{e^{w_1 x_i + c_i}}{ \sum_{j=1}^{T} e^{w_1 x_j + c_j}}$. 

\section{Learning Framework for SSG}
\label{sec:pacLearning}

\begin{table}[t]
\footnotesize
\begin{tabular}{l| l}
Notation & Meaning\\
\hline
$T, K$ & Number of targets, defender resources \\
$d_{l_p}(o, o')$ & $l_p$ distance between points $o,o'$ \\
$\bar{d_{l_p}}(o, o')$ & Average $l_p$ distance: $=d_{l_p}(o, o')/n$ \\
$X$ & Instance space (defender mixed strategies) \\
$Y$ & Outcome space (attacked target) \\
$A$ & Decision space \\
$h \in \mathcal{H}$ & $h:X \rightarrow A$ is the hypothesis function \\
$\mathcal{N}(\epsilon, \mathcal{H}, d)$ & $\epsilon$-cover of set $\mathcal{H}$ using distance $d$\\
$\mathcal{C}(\epsilon, \mathcal{H}, d)$ & capacity of $\mathcal{H}$ using distance $d$\\
$r_h(p), \hat{r}_h(\vec{z})$ & true risk, empirical risk of hypothesis $h$\\
$d_{L^1(P, d)} (f, g)$ & $L_1$ distance between functions $f,g$\\
$q^p(x)$ & parameters of true attack distribution\\
$q^h(x)$ & parameters of attack distr. predicted by $h$\\
\end{tabular}
\caption{Notations}
\label{Table1}
\vspace{-10pt}
\end{table}

First, we introduce some notations: given two $n$-dimensional points $o$ and $o'$, the $l_p$ distance $d_{l_p}$ between the two points is: $d_{l_p}(o,o') =||o - o'||_p = (\sum_{i=1}^n |o_i - o'_i|^p)^{1/p}$. In particular, $d_{l_\infty}(o,o') = ||o - o'||_\infty = \max_i |o_i - o'_i|$. Also, $\bar{d_{l_p}} =  d_{l_p}/n$.
$KL$ denotes the Kullback-Leibler divergence~\cite{Kullback59}.

We use the learning framework of Haussler~\shortcite{Haussler1992}, which includes an \emph{instance space} $X$ and \emph{outcome space} $Y$. In our context, $X$ is same as the space of defender mixed strategies $x\in X$. \emph{Outcome space} $Y$ is defined as the space of all possible categorical choices over a set of $T$  targets (i.e., choice of target to attack) for the adversary. 
Let $\attackletter_i$ denote the attacker's choice to attack the $i^{th}$ target. More formally $\attackletter_i = \langle \attackletter^1_i, \ldots, \attackletter^T_i \rangle$, where $\attackletter^j_i =1$ for $j=i$ and  otherwise $0$. Thus, $Y = \{\attackletter_1, \ldots, \attackletter_T\}$. We will use $y$ to denote any general element of $Y$. To give an example, given three targets $T_1$, $T_2$ and $T_3$, $Y=\{\attackletter_1, \attackletter_2, \attackletter_3\} = \{\langle 1,0,0 \rangle, \langle 0,1,0 \rangle, \langle 0,0,1 \rangle\}$, where $\attackletter_1$ denotes $\langle 1,0,0 \rangle$, i.e., it denotes that $T_1$ was attacked while $T_2$ and $T_3$ were not attacked, and so on. The training data are samples drawn from $Z = X\times Y$ using an unknown probability distribution, say given by density $p(x,y)$. Each training data point $(x, y)$ denotes the adversary's response $y \in Y$ (e.g., t1 or attack on target 1) to a particular defender mixed strategy $x \in X$. The density $p$ also determines the true attacker behavior $q^p(x)$ which stands for the conditional probabilities of the attacker attacking a target given $x$ so that $q^p(x) = \langle q^p_1(x), \ldots, q^p_T(x) \rangle$, where $q^p_i(x) =  p(\attackletter_i | x)$.

Haussler~\shortcite{Haussler1992} also defines a \emph{decision space} $A$, 
a \emph{space of hypothesis} (functions) $\mathcal{H}$ with elements $h:X \rightarrow A$ and a \emph{loss function} $l:Y \times A \rightarrow \mathbb{R}$. The hypothesis $h$ outputs values in $A$ that enables computing (probabilistic) predictions of the actual outcome. The loss function $l$ captures the loss when the real outcome is $y \in Y$ and the prediction of possible outcomes happens using $a \in A$.

\textbf{Example 1: Generalized SUQR} For the parametric representation of generalized SUQR in the previous section and considering our 3-target example above, $\mathcal{H}$ contains vector valued functions with $(T-1)=2$ components that form the exponents of the numerator of prediction probabilities $q_1$ and $q_2$. $\mathcal{H}$ contains two components, since the third component $q_3$ is proportional to $e^0$ as discussed above. That is, $\mathcal{H}$ contains functions of the form: $\langle w_1 (x_1 - x_3) + c_{13}, w_1 (x_2 - x_3) + c_{23} \rangle $; $\forall x\in X$. Also, $A$ is the range of the functions in $\mathcal{H}$, 
i.e., $A \subset \mathbb{R}^2$. Then, given $h(x) = \langle a_1, a_2 \rangle$, the prediction probabilities $q^h_1(x), q^h_2(x), q^h_3(x)$ are given by $q^h_i(x) = \frac{e^{a_i}}{1 + e^{a_1} + e^{a_2}}$ (assume $a_3 = 0$).

\textbf{PAC learnability:} The learning algorithm aims to learn a $h \in \mathcal{H}$ that minimizes the \emph{true risk} of using the hypothesis $h$. 
The \emph{true risk} $\textstyle r_h (p)$ of a particular hypothesis (predictor) $h$, given density function $p(x,y)$ over $Z = X\times Y$, 
is the expected loss of predicting $h(x)$ when the true outcome is $y$:
$$ r_h (p) = \int p(x,y) l(y,h(x)) \, dx \, dy $$
Of course, as $p$ is unknown the true risk cannot be computed. However, given (enough) samples from $p$, the true risk can be estimated by the \emph{empirical risk}.
The \emph{empirical risk} $\hat{r}_h(\vec{z})$, where $\vec{z}$ is a sequence of $m$ training samples from $Z$, is defined as: $\hat{r}_h(\vec{z}) = 1/m \sum_{i=1}^m l(y_i, h(x_i))$. Let $h^*$ be the hypothesis that minimizes the true risk, i.e., $r_{h^*}(p) = \inf \{r_h(p) ~|~ h \in \mathcal{H}\}$ and let $\hat{h}^*$ be the hypothesis that minimizes the empirical risk, i.e., $\hat{r}_{\hat{h}^*}(\vec{z}) = \inf \{\hat{r}_h(\vec{z}) ~|~ h \in \mathcal{H}\}$. The following is the well-known PAC learning result~\cite{Anthony2009NNL} for any \emph{empirical risk minimizing} (ERM) algorithm $\mathcal{A}$ yielding hypothesis $\mathcal{A}(\vec{z})$:
$$
\begin{array}{c}
\mbox{If }Pr(\forall h \in \mathcal{H}. |\hat{r}_h(\vec{z}) - r_h(p)| < \alpha/3 ) > 1 - \delta/2\\
\mbox{and } Pr(|\hat{r}_{\mathcal{A}(\vec{z})}(\vec{z}) - \hat{r}_{\hat{h}^*}(\vec{z})| < \alpha/3 ) > 1 - \delta/2\\
\mbox{then } Pr( |r_{\mathcal{A}(\vec{z})}(p) - r_{h^*}(p)| < \alpha ) > 1 - \delta
\end{array}
$$
The final result states that output hypothesis $\mathcal{A}(\vec{z})$ has true risk $\alpha$-close to the lowest true risk in $\mathcal{H}$ attained by $h^*$ with high probability $1 - \delta$ over the choice of training samples.
The first pre-condition states that it must be the case that for all $h \in \mathcal{H}$ the difference between empirical risk and true risk is $\frac{\alpha}{3}$-close with high probability $1 - \frac{\delta}{2}$. The second pre-condition states that the output $\mathcal{A}(\vec{z})$  of the ERM algorithm $\mathcal{A}$ should have empirical risk  $\frac{\alpha}{3}$-close to the lowest empirical risk of $\hat{h}^*$ with high probability $1 - \frac{\delta}{2}$.  A hypothesis class $\mathcal{H}$ is called $(\alpha, \delta)$-PAC learnable if there exists an ERM algorithm $\mathcal{A}$ such that $\mathcal{H}$ and $\mathcal{A}$ satisfy the two pre-conditions. In this work, our empirical risk minimizing algorithms find $\hat{h}^*$ exactly (upto precision of convex solvers, see Section~\ref{sec:ERM}), thus,  satisfying the second pre-condition; hence, we will focus more on the first pre-condition. As the empirical risk estimate gets better with increasing samples, a minimum number of samples are required to ensure that the first pre-condition holds (see Theorem~\ref{PACbound}). Hence we can relate $(\alpha, \delta)$-PAC learnability to the number of samples.


\medskip 
\noindent\textbf{Modeling security games:} Having given an example for generalized SUQR, we systematically model learning of adversary behavior in SSGs using the PAC framework for any hypothesis class $\mathcal{H}$. We assume certain properties of functions $h \in \mathcal{H}$ that we present below. First, the vector valued function $h \in \mathcal{H}$ takes the form $$h(x) = \langle h_1(x) , \ldots, h_{T-1}(x) \rangle.$$
Thus, $A$ is the product space $A_1 \times \ldots, A_{T-1}$.
Each $h_i(x)$ is assumed to take values between $[-\frac{M}{2}, \frac{M}{2}]$, where $M >\!>1$, which implies $A_i = [-\frac{M}{2}, \frac{M}{2}]$. The prediction probabilities induced by any $h$ is
$
q^h(x) = \langle q^h_1(x), \ldots, q^h_T(x)  \rangle
 $, where $q^h_i(x) = \frac{e^{h_i(x)}}{1 + \sum_i e^{h_i(x)}}$ (assume $h_T(x) = 0$). Next, we specify two classes of functions that we analyze in later sections. We choose these two functions classes because (1) the first function class represents the widely used SUQR model in literature~\cite{Nguyen13analyzingthe,Yang20142615805} and (2) the second function class is very flexible as it capture a wide range of functions and only imposes minimal Lipschitzness constraints to ensure that the functions are well behaved (e.g., continuous).


\textit{Parametric $\mathcal{H}$:} In this approach we model generalized SUQR. Generalizing from Example 1, the functions $h \in \mathcal{H}$ take a parametric form where each component function is $h_i(x) = w_{1} x_{iT} 
+ c_{iT}$. 

\textit{Non-parametric Lipschitz (NPL) $\mathcal{H}$:} Here, the only restriction we impose on functions $h \in \mathcal{H}$ is that each component function $h_i$ is $L$-Lipschitz where $L \leq \hat{K}$, for given and fixed constant $\hat{K}$. We show later (Lemma~\ref{nonparamLip}) that this implies that $q^h$ is Lipschitz also. 

Next, given the stochastic nature of the adversary's attacks, we use a loss function (same for parametric and NPL) such that minimizing the empirical risk is equivalent to maximizing the likelihood of seeing the attack data.  The loss function $l:Y \times A \rightarrow \mathbb{R}$ for actual outcome $\attackletter_i$ is defined as:
\vspace{-7pt}
\begin{equation}\label{losseq}
 l(\attackletter_i, a) = -\log \big( e^{a_i}/1 + \sum_{j=1}^{T-1} e^{a_j} \big ). 
\end{equation}
It can be readily inferred that minimizing the empirical risk (recall $\hat{r}_h(\vec{z}) = 1/m \sum_{i=1}^m l(y_i, h(x_i))$) is equivalent to maximizing the log likelihood of the training data.



\section{Sample Complexity} \label{samplecomplexity}
In this section we derive the sample complexity for the parametric and NPL case, which provides an indication about the amount of data required to learn the adversary behavior. First, we present a general result about sample complexity bounds for any $\mathcal{H}$, given our loss $l$. This result relies on sample complexity results in~\cite{Haussler1992}. The bound depends on the capacity $\mathcal{C}$ of $\mathcal{H}$, which we define after the theorem. The bound also assumes an ERM algorithm which we present for our models in Section~\ref{sec:ERM}. 
\begin{theorem} \label{PACbound}
Assume that the hypothesis space $\mathcal{H}$ is permissible\footnote{As noted in Haussler: ``This is a measurability condition defined in Pollard (1984) which need not concern us in
practice.''}. Let the data be generated by $m$
independent draws from $X \times Y$ according to $p$. Then, assuming existence of an ERM algorithm and given our loss $l$ defined in Eq.~\ref{losseq}, the least $m$ required to ensure $(\alpha, \delta)$-PAC learnability is (recall $\bar{d_{l_1}}$ is average $l_1$ distance)
$$
 \frac{576M^2}{\alpha^2} \Big( \log \frac{1}{\delta} + \log \big(8\mathcal{C}(\frac{\alpha}{96T}, \mathcal{H}, \bar{d_{l_1}}) \big) \Big)
$$
\end{theorem}
\begin{proof}[Sketch] Haussler~\cite{Haussler1992} present a result of the above form using a general distance metric defined on the space $A$ for any loss function $l$:
\vspace{-5pt}
$$
\rho(a, b) = \max_{y \in Y} |l(y, a) - l(y, b)|
$$
The main effort in this proof is relating $\rho$ to $\bar{d_{l_1}}$ for our choice of the loss function $l$ given by Equation~\ref{losseq}. We are able to show that $\rho(a,b) \leq 2T \bar{d_{l_1}} (a, b)$ for our loss function. Our result then follows from this relation (details in Appendix).
\end{proof}

The above sample complexity result is stated in terms of the capacity $\mathcal{C}(\alpha/96T, \mathcal{H}, \bar{d_{l_1}})$. Thus, in order to obtain the sample complexity of the generalized SUQR and NPL function spaces we need to compute the capacity of these function spaces. Therefore, in the rest of this section we will 
focus on computing capacity $\mathcal{C}(\alpha/96T, \mathcal{H}, \bar{d_{l_1}})$ for both the generalized SUQR and NPL hypothesis
space. First, we need to define capacity $\mathcal{C}$ of functions spaces, for which we start by defining the covering number $\mathcal{N}$ of function spaces. 
Let $d$ be a pseudo metric for the set $\mathcal{H}$. For any $\epsilon > 0$, an $\epsilon$-cover for $\mathcal{H}$
is a finite set $\mathcal{F} \subseteq \mathcal{H}$  such that for any $h \in \mathcal{H}$
there is a $f \in \mathcal{F}$ with $d(f, h) \leq \epsilon$, i.e., any element in $\mathcal{H}$ is at least $\epsilon$-close to some element of the cover $\mathcal{F}$.
The \emph{covering number} $\mathcal{N}(\epsilon, \mathcal{H}, d)$ denotes the size of the
smallest $\epsilon$-cover for set $\mathcal{H}$ (for the pseudo metric $d$). 
We now proceed to define a pseudo metric  $d_{L^1(P, d)}$ on $\mathcal{H}$ with respect to  any probability measure $P$ on $X$ and any given pseudo-metric $d$ on $A$. This pseudo-metric is the expected (over $P$) distance (with $d$) between the output of $f$ and $g$.
$$
d_{L^1(P, d)} (f, g) = \int_X d(f(x), g(x)) \; dP(x) \quad \forall f, g \in \mathcal{H}
$$
Then, $\mathcal{N}(\epsilon, \mathcal{H}, d_{L^1(P, d)})$ is the covering number for $\mathcal{H}$ for the pseudo metric $d_{L^1(P, d)}$.
However, to be more general, the capacity of function spaces provides a ``distribution-free'' notion of covering number. The \emph{capacity} $\mathcal{C}(\epsilon, \mathcal{H}, d)$ is:
$$
 \mathcal{C}(\epsilon, \mathcal{H}, d) = \sup_P \; \{\mathcal{N}(\epsilon, \mathcal{H}, d_{L^1(P, d)})\}
$$

\textit{Capacity of vector valued function:} 
The function spaces (both parametric and NPL) we consider are vector valued.
Haussler~\shortcite{Haussler1992} provides an useful technique to bound the capacity for vector valued function space $\mathcal{H}$ in terms of the capacity of each of the component real valued function space. Given $k$ functions spaces $\mathcal{H}_1, \ldots, \mathcal{H}_k$ with functions from $X$ to $A_i$, he define the \emph{free product} function space $\times_i \mathcal{H}_i$ with functions from $X$ to $A= A_1 \times \ldots A_k$ as $\times_i \mathcal{H}_i = \{\langle h_1, \ldots, h_k \rangle ~|~  h_i \in \mathcal{H}_i \}$, where $\langle h_1, \ldots, h_k \rangle (x) = \langle h_1(x), \ldots, h_k(x) \rangle$. He shows that:
\begin{equation} \label{freebound}
\mathcal{C}(\epsilon, \times_i \mathcal{H}_i, \bar{d_{l_1}}) < \prod_{i=1}^{k} \mathcal{C}(\epsilon, \mathcal{H}_i, d_{l_1})
\end{equation}
Unfortunately, a straightforward application of the above result does not give as tight bounds for capacity in the parametric case as the novel direct sum decomposition of function spaces approach we use in the sub-section. Even for the NPL case where the above result is used we still need to compute $\mathcal{C}(\epsilon, \mathcal{H}_i, d_{l_1})$ for each component function space $\mathcal{H}_i$.
\subsection{Parametric case: Generalized SUQR}
Recall that
the hypothesis function $h$ has $T-1$ component functions $w_1 x_{iT} + c_{iT}$. However, the same weight $w_1$ in all component functions implies that $\mathcal{H}$ is not a free product of component function spaces, hence we cannot use Equation~\ref{freebound} directly. However, if we consider the space of functions, say $\mathcal{H}'$, in which the $i^{th}$ component function space $\mathcal{H}'_i$ is given by $w_i x_{iT} + c_{iT}$ (note $w_i$ can be different for each $i$) then we can use Equation~\ref{freebound} to bound $\mathcal{C}(\epsilon, \mathcal{H}', \bar{d_{l_1}})$. Also, the fact that $\mathcal{H} \subset \mathcal{H}'$ allows upper bounding $\mathcal{C}(\epsilon,  \mathcal{H}, \bar{d_{l_1}})$ by $\mathcal{C}(\epsilon, \mathcal{H}', \bar{d_{l_1}})$. But, this approach results in a weaker $T \log (\frac{T}{\alpha} \log \frac{T}{\alpha})$ bound (detailed derivation using this approach is in Appendix) than the technique we use below. We obtain a $T \log (\frac{T}{\alpha})$ result below in Theorem~\ref{SUQRBound}. 

We propose a \emph{novel approach that decomposes $\mathcal{H}$ into a direct sum of two functions spaces} (defined below), each of which capture the simpler functions $w_1x_{iT}$ and $c_{iT}$ respectively. We provide a general result about such decomposition which allows us to bound $\mathcal{C}(\epsilon,  \mathcal{H}, \bar{d_{l_1}})$. We start with the following definition.
\begin{definition}
Direct-sum semi-free product of function spaces $\mathcal{G}  \subset  \times_i  \mathcal{G}_i$ and $\times_i  \mathcal{F}_i$ is defined as $\mathcal{G} \oplus \times_i  \mathcal{F}_i  = \{\langle g_1+f_1, \ldots, g_{T-1}+f_{T-1} \rangle ~|~  \langle g_1, \ldots, g_{T-1} \rangle  \in  \mathcal{G} , \langle f_1, \ldots, f_{T-1} \rangle \in \times_i  \mathcal{F}_i\}$.
\end{definition}
Applying the above definition for our case, $\mathcal{G}_i$ contains functions of the form $w x_{iT}$ ($w$ taking different values for different $g_i \in \mathcal{G}_i$). A function $\langle g_1, \ldots, g_{T-1} \rangle \in \times_i  \mathcal{G}_i$ can have different weights for each component $g_i$, and thus we consider the subset $\mathcal{G} = \{\langle g_1, \ldots, g_{T-1} \rangle ~|~ \langle g_1, \ldots, g_{T-1} \rangle \in \times_i  \mathcal{G}_i, \mbox{same coefficient $w$ for all $g_i$} \}$. $\mathcal{F}_i$ contains constant valued functions of the form $c_{i T}$ ($c_{i T}$ different for different functions $f_i \in \mathcal{F}_i$). Then, $\mathcal{H} = \mathcal{G} \oplus \times_i  \mathcal{F}_i$. 
Next, we prove a general result about direct-sum semi-free products:
\begin{lemma} \label{freeprod} If $\mathcal{H}$ is the {direct-sum semi-free product} $\mathcal{G} \oplus \times_i  \mathcal{F}_i$
$$
\mathcal{C}(\epsilon,\mathcal{H}, \bar{d_{l_1}}) <  \mathcal{C}(\epsilon/2, \mathcal{G}, \bar{d_{l_1}}) \prod_{i=1}^{T-1} \mathcal{C}(\epsilon/2, \mathcal{F}_i, d_{l_1})
$$
\end{lemma}

\begin{proof} Fix any probability distribution over $X$, say $P$. For brevity, we write $k$ instead of $T-1$.
Consider an $\epsilon/2$-cover $U_i$ for each $\mathcal{F}_i$; also let $V$ be an $\epsilon/2$-cover for $\mathcal{G}$. We claim that $V \oplus \times_i U_i $ is an $\epsilon$-cover for $\mathcal{G} \oplus \times_i  \mathcal{F}_i$. Take any function $h = \langle g_1+f_1, \ldots g_k+f_k \rangle$. Find functions $f'_i \in U_i$ such that $d_{L^1(P, d_{l_1})}(f_i, f'_i) < \epsilon/2$. Similarly, find function $g' = \langle g'_1, \ldots g'_k\rangle \in V$ such that $d_{L^1(P, \bar{d_{l_1}})}(g, g') < \epsilon/2$ where $g = \langle g_1, \ldots g_k\rangle$. Let $h' = \langle g'_1+f'_1, \ldots g'_k+f'_k \rangle$. Then,
$$
\begin{array}{l}
\displaystyle d_{L^1(P, \bar{d_{l_1}})}(h, h')  \\  
\quad = \displaystyle\int_X \frac{1}{k} \sum_{i=1}^k d_{l_1}(g_i(x) + f_i(x), g'_i(x) + f'_i(x)) \; dP(x)\\
\quad \leq \displaystyle\int_X \frac{1}{k} \sum_{i=1}^k d_{l_1}(g_i(x) , g'_i(x)) + d_{l_1}(f_i(x) , f'_i(x)) \; dP(x)\\
\quad = \displaystyle d_{L^1(P, \bar{d_{l_1}})}(g , g') + \frac{1}{k} \sum_{i=1}^k d_{L^1(P, d_{l_1})}(f_i , f'_i)\\
\quad < \epsilon/2 + \epsilon/2 = \epsilon
\end{array}
$$
Thus, the size of $\epsilon$-cover for $\mathcal{G} \oplus \times_i  \mathcal{F}_i$ is bounded by $|V|\prod_i |U_i|$. 
$$
\begin{array}{l}
\mathcal{N}(\epsilon, \mathcal{G} \oplus \times_i  \mathcal{F}_i, d_{L^1(P, \bar{d_{l_1}})}) < |V|\prod_i |U_i| \\ 
\qquad =   \mathcal{N}(\epsilon/2, \mathcal{G}, d_{L^1(P, \bar{d_{l_1}})}) \prod_{i=1}^{k} \mathcal{N}(\epsilon/2, \mathcal{F}_i, d_{L^1(P, d_{l_1})})
\end{array}
$$
Taking $\sup$ over probability distribution $P$ on both sides of the above inequality we get our desired result about capacity. 
\end{proof}

Next, we need to bound the capacity of $\mathcal{G}$ and $\mathcal{F}_i$ for our case.
We assume the range of all these functions ($g_i, f_i$) to be $[-\frac{M}{4}, \frac{M}{4}]$ (so that their sum $h_i$ lies in $[-\frac{M}{2}, \frac{M}{2}]$). We can obtain sharp bounds on the capacities of $\mathcal{G}$ and $\mathcal{F}_i$ decomposed from $\mathcal{H}$ in order to obtain sharp bounds on the overall capacity.
\begin{lemma} \label{Gbound}
 $\mathcal{C}(\epsilon, \mathcal{G}, \bar{d_{l_1}}) \leq M/4\epsilon$ and $\mathcal{C}(\epsilon, \mathcal{F}_i, d_{l_1}) \leq M/4\epsilon$.
\end{lemma} 
\begin{proof}[Sketch]
 First, note that $x_{iT} = x_i - x_T$ lies between $[-1, 1]$ due to the constraints on $x_i, x_T$.
Then, for any two functions $g, g' \in  \mathcal{G}$ we can prove following result: $d_{L^1(P, \bar{d_{l_1}})}(g, g') \leq | (w - w')|$ (details in Appendix).
 Also, note that since the range of any $g = w(x_i - x_T)$ is $[-\frac{M}{4}, \frac{M}{4}]$ and given $x_i - x_T$ lies between $[-1, 1]$, we can claim that $w$ lies between $[-\frac{M}{4}, \frac{M}{4}]$. Thus, given the distance between functions is bounded by the difference in weights, it enough to divide the $M/2$ range of the weights into intervals of size $2\epsilon$ and consider functions at the boundaries. Hence the $\epsilon$-cover has at most $M/4\epsilon$ functions.
 
 The proof for constant valued functions $\mathcal{F}_i$ is similar, since its straightforward to see the distance between two functions in this space is the difference in the constant output. Also, the constants lie in $[-\frac{M}{4}, \frac{M}{4}]$, Then, the argument is same as the $\mathcal{G}$ case.
 \end{proof}
Then, plugging the result of Lemma~\ref{Gbound} (substituting $\epsilon/2$ for $\epsilon$) into Lemma~\ref{freeprod} we obtain
$
\textstyle \mathcal{C}(\epsilon, \mathcal{H}, \bar{d_{l_1}}) < (M/2\epsilon)^T
$.
Having bounded $\mathcal{C}(\epsilon, \mathcal{H}, \bar{d_{l_1}})$, we use Theorem~\ref{PACbound} to obtain:
\begin{theorem} \label{SUQRBound}
The generalized SUQR parametric hypothesis class $\mathcal{H}$ is $(\alpha, \delta)$-PAC learnable with sample complexity\footnote{In the Appendix, we show that for standard SUQR (simpler than our generalized SUQR) the sample size is $O\big(\frac{1}{\alpha^2} ( \log\frac{1}{\delta} +  \log  \frac{T}{\alpha} )\big)$}
$$
O\Big(\big(\frac{1}{\alpha^2}\big) \big ( \log (\frac{1}{\delta}) +  T\log (\frac{T}{\alpha})  \big)\Big)
$$
\end{theorem}
The above result shows a modest $T \log T$ growth of sample complexity with increasing targets, suggesting  the parametric approach can avoid overfitting limited data with increasing number of targets; however, the simplicity of the functions captured by this approach (compared to NPL) results in lower accuracy with increasing data, as shown later in our experiments on real-world data.

\subsection{Non-Parametric Lipschitz case} 
Recall that $\mathcal{H}$ for the NPL case is defined such that each component function $h_i$ is $L$-Lipschitz  where $L \leq \hat{K}$.
Consider the functions spaces $\mathcal{H}_i$ consisting of real valued $L$-Lipschitz functions where $L \leq \hat{K}$.  Then, $\mathcal{H} = \times_i \mathcal{H}_i$. Then, using Equation~\ref{freebound}:
$
\textstyle \mathcal{C}(\epsilon, \mathcal{H}, \bar{d_{l_1}}) \leq \prod_{i=1}^{T-1} \mathcal{C}(\epsilon, \mathcal{H}_i, d_{l_1})
$.

Next, our task is to bound $\mathcal{C}(\epsilon, \mathcal{H}_i, d_{l_1})$. Consider the sup-distance metric between real valued functions: $d_{l_\infty}(h_i, h_i') = \sup_X |h_i(x) - h_i'(x)|$ for $h_i, h_i' \in \mathcal{H}_i$.
Note that $d_{l_\infty}$ is independent of any probability distribution $P$, and for all functions $h_i, h_i'$ and any $P$, $d_{L^1(P, d_{l_1})}(h_i, h_i') \leq  d_{l_\infty}(h_i, h_i')$. Thus, we can infer~\cite{Haussler1992} that for all $P$, $\mathcal{N}(\epsilon, \mathcal{H}_i, d_{L^1(P, d_{l_1})}) \leq \mathcal{N}(\epsilon, \mathcal{H}_i, d_{l_\infty})$ and then taking sup over $P$ (recall  $\mathcal{C}(\epsilon, \mathcal{H}_i, d_{l_1}) = \sup_P \; \{\mathcal{N}(\epsilon, \mathcal{H}_i, d_{L^1(P, d_{l_1})})\}$) we get
\begin{equation} \label{ontheway}
 \mathcal{C}(\epsilon, \mathcal{H}_i, d_{l_1}) \leq \mathcal{N}(\epsilon, \mathcal{H}_i,  d_{l_\infty})
\end{equation}
We bound $\mathcal{N}(\epsilon, \mathcal{H}_i,  d_{l_\infty})$ in terms of covering number for $X$ (recall $X =  \{x ~|~ x \in [0,1]^T, \sum_i x_i \leq K\}$) using results from~\cite{tikhomirov1993varepsilon}.

\begin{lemma} \label{nonparamkolmogorov}
$
\mathcal{N}(\epsilon, \mathcal{H}_i, d_{l_\infty})  \leq \Big(2 \Big\lceil \frac{M}{\epsilon} \Big\rceil + 1 \Big) \cdot 2^{\mathcal{N}(\frac{\epsilon}{2\hat{K}}, X, d_{l_\infty})}
$
\end{lemma}

To use the above result, we still need to bound  $ \mathcal{N}(\epsilon, X, d_{l_\infty})$. We do so by combining two remarkable results about Eulerian number $\eulerian{T}{k}$~\cite{knuth1998art} ($k$ has to be integral). 
\begin{itemize}
\item Laplace~\shortcite{laplace}~\cite{stanley1977eulerian} discovered that the volume of $X_k = \{x | x \in [0,1]^T, k-1 \leq \sum_i x_i \leq k \}$ is $\eulerian{T}{k} / T!$. Thus, if $X_K = \cup_{k=1}^K X_k$, then $vol(X_K) = \sum^{K}_{k = 1} vol(X_k) =  \sum^{K}_{k = 1} \eulerian{T}{k} / T!$. 
\item Also, it is known~\cite{tanny1973probabilistic} that $\eulerian{T}{k} / T! = F_T(k) - F_T(k-1)$, where $F_T(x)$ is the CDF of the probability distribution of $S_T = U _1 + \ldots + U_T$ and each $U_i$ is a uniform random variable on $[0, 1)$.
\end{itemize} 
Combining these results, $vol(X_{K+1}) = F_T(K+1)$. The volume of a $l_{\infty}$ ball of radius $\epsilon$ ($l_{\infty}$ ball is a hypercube) is $(2\epsilon)^T$~\cite{wang2005volumes}. Then, the number of balls that fit tightly (aligned with the axes) and completely inside $X_{K+1}$ is bounded by $F_T(K+1)/(2\epsilon)^T$. Since $\epsilon <\!< 1$, these balls cover $X_K = X$ completely and the tight packing ensures that the center of the balls forms an $\epsilon$-cover for $X$. 
Then, bounding $F_T(K+1)$ using Bernstein's inequality about concentration of random variables we get:
\begin{lemma} \label{coveringnumber}
For $K+1 \leq 0.5T$ (recall $K <\!< T$)
$$
 \mathcal{N}(\epsilon, X, d_{l_\infty}) \leq  e^{\frac{-3T(0.5-(K+1)/T)^2}{1-(K+1)/T}} / (2\epsilon)^T  
$$
\end{lemma}

Plugging the above result into Lemma~\ref{nonparamkolmogorov} and then using that in Equation~\ref{ontheway}, we bound $\mathcal{C}(\epsilon, \mathcal{H}_i, d_{l_1})$. Finally, Equation~\ref{freebound} gives a bound on $\mathcal{C}(\epsilon, \mathcal{H}, \bar{d_{l_1}})$ that we use in Theorem~\ref{PACbound} to obtain
\begin{theorem} \label{nonparamthm}
The non-parametric hypothesis class $\mathcal{H}$ is a $(\alpha, \delta)$-PAC learnable with sample complexity 
$$
 O\Big(\big(\frac{1}{\alpha^2} \big) \big( \log (\frac{1}{\delta}) +  (\frac{T^{T+1}}{\alpha^T}) \big) \Big)
$$
\end{theorem}
The above result shows that the sample complexity for NPL grows fast with $T$ suggesting that NPL may not be the right approach to use when the number of targets is large.

\section{Learning Algorithm} \label{sec:ERM}
As stated earlier, our loss function was designed so that the learning algorithm (empirical risk minimizer in PAC framework) was same as maximizing log likelihood of data. Indeed, for 
SUQR
, the standard MLE approach can be used 
to learn the parameters (weights) and has been used in literature~\cite{Nguyen13analyzingthe}.
However, for 
NPL, which has no parameters, maximizing likelihood only 
provides 
$h(x)$ for those 
mixed strategies $x$ that are 
in the training data. 

Hence we present a \emph{novel two step learning algorithm for the NPL case}. In the first step, we estimate the most likely value for $h_i(x)$ (for each $i$) for each $x$ in the training data, ensuring that for any pair $x, x'$ in the training data, $|h_i(x) - h_i(x')| \leq \hat{K} ||x - x'||_1$. In the second step, we construct the function 
$h_i$ with the least Lipschitz constant subject to the constraint that $h_i$ takes the values for the training data output by the first step.

More formally, assume the training data has $s$ unique values for $x$ 
in the training set and let these values be $x^1, \ldots, x^{s}$. Further, let there be $n_j$ distinct data points against 
$x^j$, i.e., $n_j$ attacks against mixed strategy $x^j$. Denote by $ n_{j,i}$ the number of attacks at each target $i$ when 
$x^j$ was used. Let $h_{ij}$ be the variable that stands for the estimate of value $h_i(x^j)$; $i \in \{1, \ldots, T\}$, $j \in \{1, \ldots, s\}$. Fix $h_{Tj} = 0$ for all $j$. Then, probability of attack on target $i$ against mixed strategy $x^j$ is given by $q_{ij} = \frac{e^{h_{ij}}}{\sum_i e^{h_{ij}}}$. Thus, the log likelihood of the training data is $\sum_{j=1}^{s} \sum_{i=1}^T n_{j, i} \log q_{ij}$. Let $Lip(\hat{K})$ denote the set of $L$-Lipschitz functions with $L \leq \hat{K}$. Using our assumption that $h_i \in Lip(\hat{K})$, the following optimization problem provides the most likely $h_{ij}$ :
$$
\begin{array}{r l}
\displaystyle\max_{h_{ij}} & \displaystyle\sum_{j=1}^{s} \sum_{i=1}^T n_{j,i} \log \frac{e^{h_{ij}}}{\sum_i e^{h_{ij}}} \\
\mbox{subject to} & \displaystyle\forall i, j, j', ~|h_{ij} - h_{ij'}| \leq \hat{K}||x^j - x^{j'}||_1 \\
 & \displaystyle\forall i, j, ~ -M/2 \leq h_{ij} \leq M/2
\end{array}
$$

Given solution $h_{ij}^*$ to the above problem, we wish to construct the solution $h_i$ such that its Lipschitz constant (given by $K_{h_i}$) is the lowest possible subject to $h_i$ taking the value $h_{ij}^*$ for $x^j$. Such a construction provides the most smoothly varying solution given the training data, i.e., we do not assume any more sharp changes in the adversary response than what the training data provides.
$$
\min_{h_i \in Lip(\hat{K})} K_{h_i} \; \mbox{ subject to } \; \forall i,j.~h_i(x^j) = h_{ij}^* \quad (\sf{MinLip})
$$
The above optimization is impractical to solve computationally as uncountably many constraints are required to relate $K_{h_i}$ to $h_i$, Fortunately, we obtain an analytical solution:
\begin{lemma} \label{MinLipSolution} The following is a solution for problem $\sf{MinLip}$
$$
 h_i(x) = \textstyle \min_j \{h_{ij}^* + K^*_i||x - x^j||_1\} 
$$
where $K^*_i = \max_{j, j': j\neq j'} \frac{| h_{ij}^* -  h_{ij'}^*|}{||x^j - x^{j'}||_1}$
\end{lemma}
\begin{proof}[Sketch]
Observe that due to the definition of $K^*$ any solution to $\sf{MinLip}$ will have Lipschitz constant $\geq K^*$. Thus, it suffices to show that the Lipschitz constant of $h_i$ is $K^*$, to prove that $h_i$ is a solution of $\sf{MinLip}$, which we show in the Appendix.
\end{proof}
Note that for any point $x^j$ is the training data we have $h_i(x^j) = h_{ij}^*$. Then the value of $h_i(x)$ for a $x$ not in the training set and close to $x^j$ is quite likely be the $h_i(x^j)$ plus the scaled distance $K_i^*||x - x^j||_1$ showing the value of $x$ is influenced by nearby training points.

\section{Utility Bounds} \label{utilitybounds}
Next, we bound the difference between the optimal utility and the utility derived from planning using the learned $h$. The utility bound is same for the parametric and NPL case. Recall that the defender receives the utility $x U q^p(x)$ when playing strategy $x$. We need to bound the difference between the true distribution $q^p(x)$ and the predicted distribution $q^h(x)$ of attacks in order to start analyzing bounds on utility.
Thus, we transform the PAC learning guarantee about the risk of output $h$ to a bound on $||q^p(x) - q^h(x)||_1$. As the PAC guarantee only bounds the risk between predicted $h$ and the best hypothesis $h^*$ in $\mathcal{H}$, in order to relate the true distribution $q^p$ and predicted distribution $q^h$, the lemma below assumes a bounded KL divergence between the distribution of the best hypothesis $q^{h^*}$ and the true distribution $q^p$.
 \begin{lemma} \label{paramrisk}
Assume $E[\mbox{KL}( q^p(x) ~||~ q^{h^*}(x))] \leq \epsilon^*$. Given an ERM $\mathcal{A}$ with output $h =\mathcal{A}(\vec{z}) $ and guarantee $Pr( |r_{h}(p) - r_{h^*}(p)| < \alpha ) > 1 - \delta$, with prob. $\geq 1 - \delta$ over training samples $\vec{z}$ we have
$$
 Pr(||q^p(x) - q^h(x)||_1 \leq \sqrt{2} \Delta) \geq 1 - \Delta
$$
where $\Delta = (\alpha + \epsilon^*)^{1/3}$ and $x$ is sampled using density $p$.
 \end{lemma}
\textbf{Utility bound: }
Next, we provide an utility bound, given the above guarantee about learned $h$.
Let the optimal strategy computed using the learned adversary model $h$ be $\tilde{x}$, i.e., $\tilde{x}^T U q^h(\tilde{x}) \geq x'^T U q^h (x')$ for all $x'$. Let the true optimal defender mixed strategy be $x^*$ (optimal 
w.r.t. true attack distribution $q^p(x)$), so that the maximum defender utility is $x^{*T} U q^p(x^*)$. Let $B(x, \epsilon)$ denote the $l_1$ ball of radius $\epsilon$ around $x$.
We make the following assumptions: 
\begin{enumerate}
\item $h_i$ is $\hat{K}$-Lipschitz $\forall i$ and $q^p$ is $K$-Lipschitz in $l_1$ norm.
\item $\exists \mbox{ small }\epsilon$ such that $Pr(x \in B(x^*, \epsilon)) > \Delta$ over choice of $x$ using $p$.
\item $\exists  \mbox{ small } \epsilon$ such that  $Pr(x \in B(\tilde{x}, \epsilon)) > \Delta $ over choice of $x$ using $p$.
\end{enumerate}
While the first assumption is mild~\footnote{Lipschitzness is a mild restriction on function classes.}, the last two assumptions for small $\epsilon$ mean that the points $x^*$ and $\tilde{x}$ must not lie in low density regions of the distribution $p$ used to sample the data points. In other words, there should be many defender mixed strategies in the data of defender-adversary interaction that lie near $x^*$ and $\tilde{x}$. We discuss the assumptions in details after the technical results below.
Given these assumptions, we need Lemma 7 that relates assumption (1) to Lipschitzness of $q^h$ in order to obtain the utility bound.
\begin{lemma} \label{nonparamLip} If $h_i$ is $\hat{K}$-Lipschitz then 
$\forall x, x' \in X.~||q^h(x) - q^h(x')||_1 \leq 3 \hat{K} ||x - x'||_1$, i.e., $q^h(x)$ is $3\hat{K}$-Lipschitz.
\end{lemma}
Then, we can prove the following:
\begin{theorem} \label{paramutilitybound}
Given above assumptions and the results of Lemma~\ref{paramrisk} and~\ref{nonparamLip}, with prob. $\geq 1- \delta$ over the training samples the expected utility $\tilde{x}^T U q^h(\tilde{x})$ for the learned $h$ is at least 
$$
 x^{*T} U q^p(x^*) - (K+1)\epsilon -  2\sqrt{2}\Delta - 6\hat{K} \epsilon 
$$
\end{theorem}

\textbf{Discussion of assumptions}: A puzzling phenomenon observed in recent work on learning in SSGs is that good prediction accuracy of the learned adversary behavior is not a reliable indicator of the defender's performance in practice~\cite{Ford15a}.
The additional assumptions, over and above the PAC learning guarantee, are made to bound the utility deviation from the optimal utility point towards the possibility of such occurrences. Recall that the second assumption requires the existence of many defender mixed strategies in the dataset near the utility optimal 
strategy $x^*$. Of course $x^*$ is not known apriori, hence in order to guarantee utility close to the highest possible utility the dataset must contain defender mixed strategies from all regions of the 
mixed strategy space; or at-least if it is known that some regions of the mixed strategies dominate other parts in terms of utility then it is enough to have mixed strategies from these regions. Thus, following our assumption, better utility can be achieved by collecting attack data against a variety of mixed strategies rather than many attacks against few mixed strategies. 

Going further, we illustrate with a somewhat extreme example where violating our assumptions can lead to this undesirable phenomenon.
For the purpose of illustration, consider the extreme example where probability distribution $p$ (recall data points are sampled using $p$) puts all probability mass on $x_0$, where the the utility for $x_0$ is much lower than $x^*$. Hence, the dataset will contain only one defender mixed strategy $x_0$ (with many attacks against it). Due to Lipschitzness (assumption 1), the large utility difference between $x_0$ and $x^*$ implies that $x_0$ is not close to $x^*$ which in turn violates assumption 2. This example provides a very good PAC guarantee since there is no requirement for the learning algorithm to predict accurately for any other mixed strategies (which occur with zero probability) in order to have good prediction accuracy. The learning technique needs to predict well only for $x_0$ to achieve a low $\alpha, \delta$.  As a result the defender strategy computed against the learned adversary model may not be utility maximizing because of the poor predictions for all defender mixed strategies other than the low utility yielding $x_0$. More generally, good prediction accuracy can be achieved by good predictions only for the mixed strategies that occur with high probability.

Indeed, in general, the prediction accuracy in the PAC model (and any applied machine learning approach) is not a reliable indicator of good prediction over the entire space of defender mixed strategies unless, following our assumption 2, the dataset has attacks against strategies from all parts of the mixed strategy space. However, in past work~\cite{Dehghani15a,cui2014empirical} researchers have focused on gathering a lot of attack data but on limited number of defender strategies. We believe that our analysis, in addition to providing a principled explanation of prior observations, provides guidance towards methods of discovering the defender's utility maximizing strategy.

\begin{figure}[!t]
    \centering
   \vspace{-2pt}
	\subfigure[SUQR, Uganda data, coarse-grained prediction]{
    \label{fig:Uganda_param_2012_cg}
    \includegraphics[width = 0.46\linewidth, height=1.8cm]{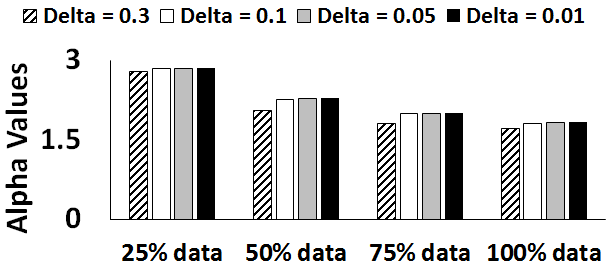}
    }	
	\subfigure[NPL, Uganda data, coarse-grained prediction]{
    \label{fig:Uganda_nonparam_2012_cg}
    \includegraphics[width = 0.46\linewidth, height=1.8cm]{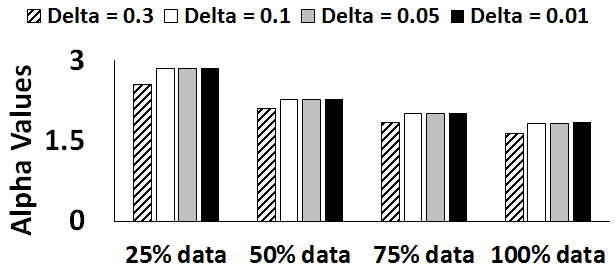}
    }	
	\subfigure[SUQR, Uganda data, fine-grained prediction]{
    \label{fig:Uganda_param_2012_fg}
    \includegraphics[width = 0.46\linewidth, height=1.8cm]{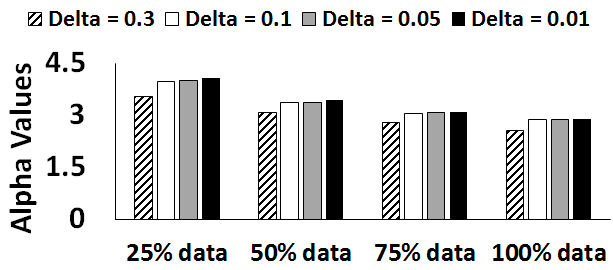}
    }	
	\subfigure[NPL, Uganda data, fine-grained prediction]{
    \label{fig:Uganda_nonparam_2012_fg}
    \includegraphics[width = 0.46\linewidth, height=1.8cm]{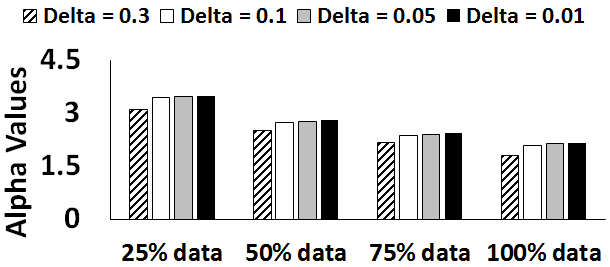}
    }	
	\subfigure[Generalized SUQR, Uganda Data, fine-grained prediction]{
    \label{fig:Uganda_Ourparam_2012_fg}
    \includegraphics[width = 0.46\linewidth, height=1.8cm]{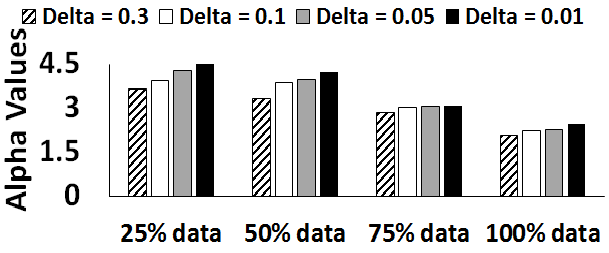}
    }	
    \subfigure[Generalized SUQR, Uganda Data, coarse-grained prediction]{
    \label{fig:Uganda_Ourparam_2012_cg}
    \includegraphics[width = 0.46\linewidth, height=1.8cm]{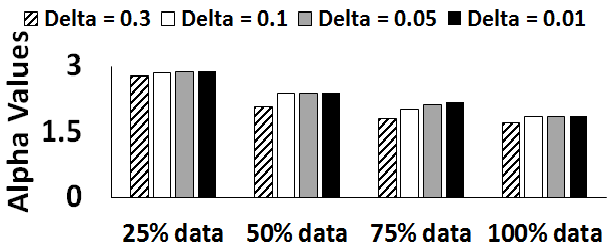}
    }
    \subfigure[Parametric, AMT data]{
    \label{fig:AMT_param_payoff1}
    \includegraphics[width = 0.46\linewidth, height=1.8cm]{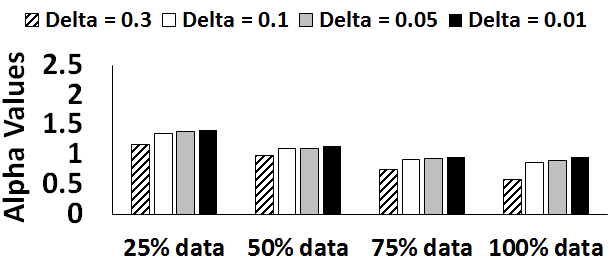}
    }	
	\subfigure[NPL, AMT data]{
    \label{fig:AMT_nonparam_payoff1}
    \includegraphics[width = 0.46\linewidth, height=1.8cm]{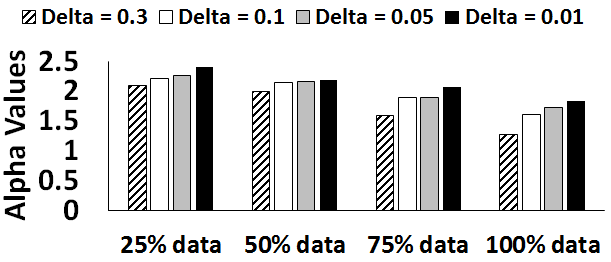}
    }	
    \caption{\footnotesize{Results on Uganda and AMT datasets for both the parametric and NPL learning settings.}}
    \vspace{-5pt}
\end{figure}


\section{Experimental Results}
We show experimental results on two datasets: (i) real-world poaching data from QENP (obtained from \cite{Nguyen15a}); (ii) data from human subjects experiments on AMT (obtained from \cite{Kar15a}), to estimate prediction errors and the amount of data required to reduce the error for both the parametric and NPL learning settings. Also, we compare the NPL approach with both the standard as well as the generalized SUQR approach and show that: (i) the NPL approach, while computationally slow, outperforms the standard SUQR model 
for Uganda data; and (ii) the performance of generalized SUQR is in between NPL and standard SUQR.

For each dataset, we conduct four experiments with 25\%, 50\%, 75\% and 100\% of the original data. We create 100 train-test splits in each of the four experiments per dataset. For each train-test split we compute the average prediction error $\alpha$ (average difference between the log-likelihoods of the attacks in the test data using predicted and actual attack probabilities). We report the $\alpha$ value at the $1-\delta$ percentile of the 100 $\alpha$ values, e.g., reported $\alpha = 2.09$ for $\delta=0.1$ means that 90 of the 100 test splits have $\alpha <2.09$. 

\subsection{Real-World Poaching data}
We first present results of our experiments with real-world poaching data. 
The dataset obtained contained information about features such as ranger patrols and animal densities, which are used as features in our SUQR model, and the poachers' attacks, with 40,611 total observations  recorded by rangers at various locations in the park. The park area was discretized into 2423 grid cells, with each grid cell corresponding to a 1km$^2$ area within the park. After discretization, each observation fell within one of 2423 target cells and the animal densities and the number of poaching attacks within each target cell were then aggregated. The dataset contained 655 poachers' attacks in response to the defender's strategy for 2012 at QENP. Although the data is reliable because the rangers recorded the latitudes and longitudes of the location for each observation using a GPS device, it is important to note that this data set is extremely noisy because of: (i) Missing observations: all the poaching events are not recorded because the limited number of rangers cannot patrol all the areas in this park all the time; (ii) Uncertain feature values: the animal density feature is also based on incomplete observations of animals; (iii) Uncertain defender strategy: the actual defender mixed strategy is unknown, and hence, we estimate the mixed strategies based on the provided patrol data.

In this paper, we provide two types of prediction in our experiments: (i) fine-grained and (ii) coarse-grained. First, to provide a baseline for our error measures, we use the same coarse-grained prediction approach as reported by Nguyen et al. \cite{Nguyen15a}, in which the authors only predict whether a target will be attacked or not. The results for coarse-grained predictions with our performance metric ($\alpha$ values for different $\delta$) are shown in Figs. \ref{fig:Uganda_param_2012_cg}, \ref{fig:Uganda_nonparam_2012_cg} and \ref{fig:Uganda_Ourparam_2012_cg}. Next, in the fine-grained prediction approach we predict the \textit{actual} number of attacks on each target in our test set; these results are shown in Figs. \ref{fig:Uganda_param_2012_fg}, \ref{fig:Uganda_nonparam_2012_fg} and \ref{fig:Uganda_Ourparam_2012_fg}. In \cite{Nguyen15a}, the authors used a particular metric for prediction performance called the area under the ROC curve (AUC), which we will discuss later in the section. 



From our fine-grained and coarse-grained prediction approaches, we make several important observations. First, we observe that 
$\alpha$ decreases with increasing sample size at a rate proportional to 
$\frac{1}{\sqrt m}$, where $m$ is the number of samples. 
For example, in Fig. \ref{fig:Uganda_param_2012_cg}, the $\alpha$ values corresponding to $\delta=0.01$ (black bar) for the 25\%, 50\%, 75\% and 100\% data are 2.81, 2.38, 2.18 and 1.8 respectively, which fits $\frac{1}{\sqrt m}$with a goodness-of-fit, i.e., $r^2$=0.97. This observation supports the relationship between $\alpha$ and $m$ shown in Theorem 3 and can be used to approximately infer the number of samples required to reduce the prediction error to a certain value. For example, assuming we collect same number of samples (=655) per year, to reduce $\alpha$ from 1.8 to 1.64, we would require two more years of data. Note here that $\alpha$ is in log-scale and hence the decrease is significant. It is also worth noting here that, for a random classifier, 
we observe a value of $\alpha=6.3$ for $\delta=0.01$ while performing coarse grained predictions with 25\% data. This is more than $\alpha=2.81$ for $\delta=0.01$ obtained for our standardized SUQR model while performing coarse grained predictions with 25\% data. As $\alpha$ is in the log-scale, the increase in error is actually more than two-fold.

Our second observation is that $\alpha$ values for fine-grained predictions (e.g., 2.9 for $\delta=0.01$ and 100\% data for the standardized SUQR model in Fig. \ref{fig:Uganda_param_2012_fg}) are understandably higher than the corresponding values for coarse-grained predictions (1.8 for $\delta=0.01$ and 100\% data for SUQR in Fig. \ref{fig:Uganda_param_2012_cg}) because in the fine-grained case we predict the exact number of attacks. 

Third, we observe that the performance of the generalized SUQR model (e.g., 2.47 for $\delta=0.01$ and 100\% data in Fig. \ref{fig:Uganda_Ourparam_2012_fg}) is better in most cases than that of the standardized SUQR approach (2.9 for $\delta=0.01$ and 100\% data in Fig. \ref{fig:Uganda_param_2012_fg}), but worse than the NPL approach (2.15 for $\delta=0.01$ and 100\% data in Fig. \ref{fig:Uganda_nonparam_2012_fg}).

Finally, we observe that our NPL model performs better than its parametric counterparts in predicting future poaching attacks for the fine-grained case (see example in previous paragraph), indicating that the true adversary behavior model may indeed be more complicated than what can be captured by SUQR. 


\textbf{Relation to previous work:} In earlier work, Nguyen et al. \cite{Nguyen15a} uses the area under curve (of a ROC curve) metric to demonstrate the performance of their approaches. 
The AUC value of 0.73 reported in their paper is an alternate view of our $\alpha, \delta$ metric for the coarse grained prediction approach. 
While there has been alternate analysis in terms of measuring prediction performances with the AUC metric in earlier papers, in our paper we have shown new trends and insights with the $\alpha, \delta$ metric through analysis from the PAC model perspective, which is missing in earlier work. We show: (i) sample complexity results and the relationship between increasing number of samples and the reduction in prediction error for each of our models; (ii) the differences in errors while learning a vector valued response function (fine-grained prediction) as opposed to classifying targets as attacked or not (coarse-grained prediction); and (iii) comparison of the performance of our new NPL model with other parametric approaches in terms of both fine-grained and coarse-grained predictions and its effectiveness on real-world poaching data which was not shown in previous work. 

\begin{table}[t]
\footnotesize
\centering
\begin{tabular}{|p{1.8cm}|p{1.5cm}|p{1.8cm}|p{1.5cm}|}
\hline
{Uganda Parametric} & {Uganda NPL} & {AMT~~~ Parametric} & {AMT NPL}\tabularnewline
\hline
{0.7188} & {121.24} & {0.91} & {123.4}\tabularnewline
\hline
\end{tabular}
\caption{\footnotesize{Runtime results (in secs.) for one train-test split}}
\label{Table2}
\vspace{-15pt}
\end{table}

\subsection{AMT data} Here we show fine-grained prediction results on real-world AMT data obtained from \cite{Kar15a} 
to demonstrate the performance of both our approaches on somewhat cleaner data. This dataset is cleaner than the Uganda data because: (i) all attacks are observed, and (ii) animal densities and deployed defender strategies are known. 
The dataset consisted of 16 unique mixed strategies.
There were an average of 40 attack data points per mixed strategy. Each attack had been conducted by an unique individual recruited on AMT. 
We used attack data corresponding to 11 randomly chosen mixed strategies for training and data for the remaining mixed strategies for testing. 
Results 
are shown in Figs. \ref{fig:AMT_param_payoff1} 
and \ref{fig:AMT_nonparam_payoff1}. 
We observe that: (i) $\alpha$ values in this case are lower as compared to the Uganda data as the AMT data is cleaner; and (ii) the NPL model's performance on this dataset is poor as compared to SUQR due to, (a) low number of samples in AMT data, and (b) real-world poacher behavior may be more complicated than that of AMT participants and hence SUQR in this case was able to better capture AMT participants' behavior with limited number of samples.\footnote{In the Appendix we show that, on simulated data, $\alpha$ does indeed approach zero and NPL outperforms SUQR with enough samples.} 


\textbf{Runtime:} While running on Matlab R2015a on an Intel Core i7-5500 CPU@2.40Ghz, 8GB RAM machine with a 64-bit Windows 10, on average, the NPL computation takes longer than the parametric setting, as shown in Table \ref{Table2}. 

\section{Conclusion}

Over the last couple of years, a lot of work has used learning methods to learn bounded rationality adversary behavioral model, but there has been no formal study of the learning process and its implication on the defender's performance. The lack of formal analysis also means that many practical questions go unanswered. We have advanced the state of the art in learning of adversary behaviors in SSGs, in terms of their analysis and implications of such learned behaviors on defender's performance.

While we used the PAC framework, it is not an out of the box approach. We needed
to come up with innovative techniques to obtain sharp bounds for our case. Furthermore, we also provide a new non-parametric learning approach which showed promising results with real-world data. Finally, we provided a principled explanation of observed phenomenon that prediction accuracy is not enough to guarantee good defender performance. We explained why datasets with attacks against a variety of defender mixed strategies helps in achieving good defender performance. Finally, we hope this work leads to more theoretical work in learning of adversary models and the use of non-parametric models in the real world.

\clearpage

\bibliography{References}
\bibliographystyle{abbrv}

\clearpage
\appendix

The Appendix is structured as follows: Section~\ref{app:proof} contains the missing proofs, Section~\ref{app:extension} contains the result of the applicability of our techniques for Stackelberg games, Section~\ref{app:stsuqr} constains results about the sample complexity of standard SUQR, Section~\ref{app:hausslerproof} contains the weaker sample complexity bound result for the generalized SUQR model derived using the approach of Haussler and Section~\ref{app:experiments} contains additional experiments. 

\section{Proofs} \label{app:proof}
\textbf{Proof of Theorem~\ref{PACbound}}
\begin{proof} First, Haussler uses the following pseudo metric $\rho$ on $A$ that is defined using the loss function $l$:
$$\begin{array}{c}\rho(a, b) = \max_{y \in Y} |l(y, a) - l(y, b)|. \end{array}$$ 
To start with, relying on Haussler's result, we show 
$$Pr(\forall h \in \mathcal{H}. |\hat{r}_h(\vec{z}) - r_h(p)| < \frac{\alpha}{3}) \geq 1- 4\mathcal{C}\Big(\frac{\alpha}{48}, \mathcal{H}, \rho \Big)e^{-\frac{\alpha^2m}{576M^2}}$$
Choose $\alpha = \alpha'/4M$ and $\nu = 2M$ in Theorem 9 of \cite{Haussler1992}. Using property (3) (Section 2.2, \cite{Haussler1992}) of $d_v$ we obtain $|r - s| \leq \epsilon$ whenever $d_v(r, s) \leq \alpha'$. Using this directly in Theorem 9 of Haussler~\shortcite{Haussler1992} we obtain the desired result above.

Note the dependence of the above probability on $m$ (the number of samples), and compare it to the first pre-condition in the PAC learning result. By equating $\delta/2$ to $4\mathcal{C}(\alpha/48, \mathcal{H}, \rho )e^{-\frac{\alpha^2m}{576M^2}}$, we derive the sample complexity as
$$
m \geq \frac{576M^2}{\alpha^2} \log \frac{8\mathcal{C}(\alpha/48, \mathcal{H}, \rho )}{\delta}
$$

We wish to compute a bound on $\mathcal{C}(\epsilon, \mathcal{H}, \rho)$ in order to use the above result to obtain sample complexity.  First, we prove that 
$
\rho \leq 2T d_{\bar{l_1}}$ for the loss function we use. This result is used to bound $\mathcal{C}(\epsilon, \mathcal{H}, \rho)$, since, it is readily verified from definition that $\mathcal{C}(\epsilon, \mathcal{H}, \rho) \leq \mathcal{C}(\epsilon/2T, \mathcal{H}, d_{\bar{l_1}})$. Such a bounding directly gives
$$
m \geq \frac{576M^2}{\alpha^2} \log \frac{8\mathcal{C}(\alpha/96T, \mathcal{H}, \rho )}{\delta}
$$
Below we prove that $
\rho \leq 2T d_{\bar{l_1}}$.
\begin{lemma} \label{distancelemma} Given the loss function defined above, we have
$\rho(a,b) \leq 2 \max_i |a_i - b_i| \leq 2 \sum_i |a_i - b_i| \leq 2T d_{\bar{l_1}} (a, b) $
\end{lemma}
\begin{proof}
By definition, $\rho(a,b) = \max_i \Big|-a_i + b_i + \log\frac{1 + \sum_{i=1}^{T-1} e^{a_i}}{1 + \sum_{i=1}^{T-1} e^{b_i}} \Big | \leq \max_i|a_i - b_i | + \big| \log\frac{1 + \sum_{i=1}^{T-1} e^{a_i}}{1 + \sum_{i=1}^{T-1} e^{b_i}} \Big |$.
There is $j$ and $k$ such that $max_r = \frac{e^{a_j}}{e^{b_j}} \geq \frac{e^{a_i}}{e^{b_i}}$ for all $i$ and $min_r = \frac{e^{a_k}}{e^{b_k}} \leq \frac{e^{a_i}}{e^{b_i}}$ for all $i$. Thus, 
$$\log \frac{1 + min_r t}{ 1 + t} \leq \log\frac{1 + \sum_{i=1}^{T-1} e^{a_i}}{1 + \sum_{i=1}^{T-1} e^{b_i}} \leq \log \frac{1 + max_r t}{1 +t }$$
where $t = \sum_{i=1}^{T-1} e^{b_i}$. The greatest positive value of the RHS is $\log max_r \leq |a_j - b_j|$ and least negative value possible for LHS is $\log min_r \geq -|a_k - b_k|$. Thus, 
$$ \big | \log\frac{1 + \sum_{i=1}^{T-1} e^{a_i}}{1 + \sum_{i=1}^{T-1} e^{b_i}} \big| \leq \max_i \big| a_i - b_i \big|$$
Hence, we obtain $\rho(a,b) = \max_i | l(y_i, a) - l(y_i, b)| \leq 2 \max_i |a_i - b_i|$, and the last inequality is trivial.
\end{proof}
Thus, using the above result we get
$$
m \geq \frac{576M^2}{\alpha^2} \log \frac{8\mathcal{C}(\alpha/96T, \mathcal{H}, d_{\bar{l_1}} )}{\delta}
$$
\end{proof}


\noindent\textbf{Proof of Lemma~\ref{Gbound}}
\begin{proof}
 First, note that $x_{iT} = x_i - x_T$ lies between $[-1, 1]$ due to the constraints on $x_i, x_T$.
Then, for any two functions $g, g' \in  \mathcal{G}$ we have the following result:
 $$
 \begin{array}{l}
 d_{L^1(P, \bar{d_{l_1}})}(g, g') = \\  
\quad = \displaystyle\int_X \frac{1}{T-1} \sum_{i=1}^{T-1} d_{l_1}( w(x_i - x_T),  w'(x_i - x_T) ) \; dP(x)\\
\quad = \displaystyle\int_X \frac{1}{T-1} \sum_{i=1}^{T-1} | (w - w')(x_i - x_T)| \; dP(x)\\
\quad \leq \displaystyle\int_X \frac{1}{T-1} \sum_{i=1}^{T-1} | (w - w')| \; dP(x) \, = \, | (w - w')|\\
 \end{array}
 $$
 Also, note that since the range of any $g = w(x_i - x_T)$ is $[-\frac{M}{4}, \frac{M}{4}]$ and given $x_i - x_T$ lies between $[-1, 1]$, we can claim that $w$ lies between $[-\frac{M}{4}, \frac{M}{4}]$. Thus, given the distance between functions is bounded by the difference in weights, it enough to divide the $M/2$ range of the weights into intervals of size $2\epsilon$ and consider functions at the boundaries. Hence the $\epsilon$-cover has at most $M/4\epsilon$ functions.
 
 The proof for constant valued functions $\mathcal{F}_i$ is similar, since its straightforward to see the distance between two functions in this space is the difference in the constant output. Also, the constants lie in $[-\frac{M}{4}, \frac{M}{4}]$, Then, the argument is same as the $\mathcal{G}$ case.
 \end{proof}
 
\noindent\textbf{Proof of Lemma~\ref{nonparamkolmogorov}}
\begin{proof}
First, the space of functions $\hat{\mathcal{H}} = \{h/\hat{K} ~|~ h \in \mathcal{H}_i \}$ is Lipschitz with Lipschitz constant $\leq 1$ and $|h_i(x)| \leq M/2\hat{K}$. Clearly $\mathcal{N}(\epsilon, \mathcal{H}_i, d_{l_\infty}) \leq \mathcal{N}(\epsilon/\hat{K}, \hat{\mathcal{H}}, d_{l_\infty})$. Using the following result from~\cite{tikhomirov1993varepsilon}: for any Lipschitz real valued function space $\mathcal{H}$ with constant $1$, any positive integer $s$ and any distance $d$
$$
\mathcal{N}(\epsilon, \mathcal{H}, d_{l_\infty}) \leq \Big(2\Big\lceil \frac{M(s+1)}{2\hat{K}\epsilon} \Big\rceil + 1\Big) \cdot
(s+1)^{\mathcal{N}(\frac{s\epsilon }{s+1}, X, d)}$$
Then, 
we get the bound on $\mathcal{N}(\epsilon/\hat{K}, \hat{\mathcal{H}}, d_{l_\infty})$ by choosing $s=1$ and $d = d_{l_\infty}$, and hence obtain the desired bound on $\mathcal{N}(\epsilon, \mathcal{H}_i, d_{l_\infty})$.
\end{proof}

\noindent\textbf{Proof of Lemma~\ref{coveringnumber}}
\begin{proof}
 For ease of notation, we do the proof with $k$ standing for $K+1$.
Let $Y_i = U_i - 0.5$, then
$|Y| \leq 1/2$
and
$S_T - 0.5T = \sum_i Y_i$. Using Bernstein's inequality with the fact that $E[Y_i^2] = 1/12$
$$
P(\sum_i Y_i  = S_T - 0.5T \leq - t) \leq e^{\frac{-0.5t^2}{T/12 + t/6}}
$$
Thus, 
$P(S_T \leq 0.5T - t) \leq e^{\frac{0.5t^2}{T/12 + t/6}}$.
Take $k = 0.5T - t$, and hence $t = 0.5T - k = T(0.5 - k/T)$. Hence,
$$P(S_T \leq k) \leq  e^{\frac{-3T(0.5-k/T)^2}{1-k/T}}$$
\end{proof}

\noindent\textbf{Proof of Theorem~\ref{nonparamthm}}
\begin{proof}
Given the results of Lemma~\ref{nonparamkolmogorov}, we get the sample complexity is of order
$$
\frac{1}{\alpha^2}\Big (  \log\frac{1}{\delta} + T \big(\mathcal{N}(\frac{\alpha}{T}, X, d_{l_1}) \big) \Big )
$$
Now, suing result of Lemma~\ref{coveringnumber}, we get the required order in the Theorem. We wish to note that if $K/T$ is a constant then the $O(e^{-T})$ in Lemma~\ref{coveringnumber} gets swamped by the $T^T$ term. However, in practice for fixed $T$, this term does provide lower actual complexity bound than what is indicated by the order.
\end{proof}
 
 \noindent\textbf{Proof of Lemma~\ref{MinLipSolution}}
\begin{proof}
Observe that due to the definition of $K^*$ any solution to $\sf{MinLip}$ will have Lipschitz constant $\geq K^*$. Thus, it suffices to show that the Lipschitz constant of $h_i$ is $K^*$, to prove that $h_i$ is a solution of $\sf{MinLip}$. Take any two $x, x'$. If the min in the expression for $h_i$ occurs for the same $j$ for both $x, x'$ then $|h_i(x) - h_i(x')|$ is given by $K^*|||x - x^{j}||_1 - ||x' - x^{j}||_1|$. By application of triangle inequality 
$$
-||x - x'||_1 \leq ||x - x^{j}||_1 - ||x' - x^{j}||_1 \leq ||x - x'||_1
$$
Thus, $|h_i(x) - h_i(x')| \leq K^* ||x - x'||_1$.

For the other case when the min for $x$ occurs at some $j$ and min for $x'$ at some $j'$ we have the following: $h_i(x') = h_{ij'} + K^*||x' - x^{j'}||_1$ and $h_i(x) = h_{ij} + K^*||x - x^{j'}||_1$ . Also, due to the min, $h_i(x') \leq  h_{ij} + K^*||x' - x^{j}||_1 = h_i(x) + K^*||x' - x^{j}||_1 - K^*||x - x^{j'}||_1$. Thus, we get
$$
h_i(x') - h_i(x) \leq K^*(||x' - x^{j}||_1 - ||x - x^{j}||_1) \leq K^* ||x' - x||_1
$$
Using the symmetric case inequality for $x$ we get
$$
h_i(x) - h_i(x') \leq K^*(||x - x^{j}||_1 - ||x' - x^{j}||_1) \leq K^* ||x - x'||_1
$$
Combining both these we can claim that $|h_i(x) - h_i(x')| \leq  K^* ||x' - x||_1$.
Thus, we have proved that $h_i$ is $K^*$ Lipschitz, and hence a solution of $\sf{MinLip}$.
\end{proof}
 
 \noindent\textbf{Proof of Lemma~\ref{paramrisk}}
 \begin{proof}
Let $p_X$ be the marginal of $p(x,y)$ for space $X$. Define the expected entropy $E[H(x)] = \int p_X(x)  \sum_{i=1}^T I_{y = \attackletter_i} q^p_i (x) \log q^p_i (x) \, dx$. Given the loss function, we know that $ r_h(p) = -\int p(x, y)  \sum_{i=1}^T I_{y = \attackletter_i} \log q^h_i (x) \, dx \, dy$. This is same as 
$-\int p_X(x)  \sum_{i=1}^T I_{y = \attackletter_i} q^p_i (x) \sum_{i=1}^T I_{y = \attackletter_i} \log q^h_i (x) \, dx \, dy$. This reduces to
$-\int p_X(x)  \sum_{i=1}^T I_{y = \attackletter_i} q^p_i (x) \log q^h_i (x) \, dx \, dy$.
Thus, we have
$$
E[H(x)] + r_h(p) = \int p_X(x)  \sum_{i=1}^T I_{y = \attackletter_i} q^p_i (x) \log \frac{q^p_i (x)}{q^h_i (x)} \, dx \, dy
$$
Hence, we obtain
$$
E[H(x)] + r_h(p) = E[\mbox{KL}(q^p(x) ~||~ q^h(x))]
$$Hence, $|r_h(p) - r_{h^*}(p)|$ is equal to 
$$
 |E[\mbox{KL}( q^p(x) ~||~ q^h(x))] - E[\mbox{KL}( q^p(x) ~||~ q^*(x))]|
$$ Thus, from the assumptions, we get 
$E[\mbox{KL}( q^p(x) ~||~ q^h(x))] \leq \alpha + \epsilon^*$ with probability $ \geq 1 - \delta$. Next, using Markov inequality, with probability $\geq 1 - \delta$ $$Pr(\mbox{KL}( q^p(x) ~||~ q^h(x)) \geq (\alpha + \epsilon^*)^{2/3}) \leq (\alpha + \epsilon^*)^{1/3}$$
that is using the notation $\Delta = (\alpha + \epsilon^*)^{1/3}$, with probability $\geq 1 - \delta$
$$Pr(\mbox{KL}( q^p(x) ~||~ q^h(x)) \leq \Delta^{2/3}) \geq 1 - \Delta^{1/3}$$

Using Pinkser's inequality we get $(1/2) ||q^p(x) - q^h(x)||_1^2 \leq \mbox{KL}( q^p(x) || q^h(x))$. That is, the event $\mbox{KL}( q^p(x) ~||~ q^h(x)) \leq \Delta^{2/3}$ implies the event $||q^p(x) - q^h(x)||_1 \leq \sqrt{2} \Delta$. Thus, $Pr(||q^p(x) - q^h(x)||_1 \leq \sqrt{2} \Delta) \geq Pr(\mbox{KL}( q^p(x) ~||~ q^h(x)) \leq \Delta^{2/3})$.  Thus, we obtain: with probability $\geq 1 - \delta$, 
$
Pr(||q^p(x) - q^h(x)||_1 \leq \sqrt{2} \Delta) \geq 1 - \Delta
$.

\end{proof}

\noindent\textbf{Proof of Lemma~\ref{nonparamLip}}
\begin{proof}
We know that $q^h_i(x) = \frac{e^{h_i(x)}}{\sum_ j e^{h_j(x)}}$ (assume $h_T(x) = 0$). Thus,
$$
|q^h_i(x) - q^h_i(x')| = q^h_i(x')|e^{h_i(x) - h_i(x')} \frac{\sum_j e^{h_j(x')}}{\sum_j e^{h_j(x)}} - 1|
$$

Let $r$ denote $ \frac{\sum_j e^{h_j(x')}}{\sum_j e^{h_j(x)}}$. There is $l$ and $k$ such that $max_r = \frac{e^{h_l(x')}}{e^{h_l(x)}} \geq \frac{e^{h_j(x')}}{e^{h_j(x)}}$ for all $j$ and $min_r = \frac{e^{h_k(x')}}{e^{h_k(x)}} \leq \frac{e^{h_j(x')}}{e^{h_j(x)}}$ for all $j$. Then, $min_r \leq r \leq max_r$ First, note that due to our assumption that for each $i$ $|h_i(x') - h_i(x)| \leq \hat{K} ||x' - x||_1$, we have 
$$
e^{- \hat{K} ||x' - x||_1} \leq min_r \leq r  \leq max_r \leq e^{\hat{K} ||x' - x||_1}
$$

Using the Lipschitzness we can also claim that $e^{- \hat{K} ||x' - x||_1} \leq e^{h_i(x) - h_i(x')} \leq e^{\hat{K} ||x' - x||_1}$. Thus, 
$$
e^{- 2\hat{K} ||x' - x||_1} \leq e^{h_i(x) - h_i(x')} \cdot r \leq e^{2\hat{K} ||x' - x||_1}
$$
Since, $e^{- 2\hat{K} ||x' - x||_1} < 1$ and $e^{2\hat{K} ||x' - x||_1} > 1$ we have

$$
|e^{h_i(x) - h_i(x')} r - 1| \leq \max(|e^{- 2\hat{K} ||x' - x||_1} - 1|, |e^{2\hat{K} ||x' - x||_1} - 1|)
$$

Also, it is a fact that $|e^y - 1| \leq 1.5|y|$ for $|y| \leq 3/4$. Thus, we obtain
$$
|e^{h_i(x) - h_i(x')} r - 1| \leq 3 \hat{K} ||x' - x||_1 \mbox{ for } 2\hat{K} ||x' - x||_1 \leq 3/4
$$
Thus, $||q^h(x') - q^h(x)||_1 = \sum_i |q^h_i(x) - q^h_i(x')| = \sum_i q^h_i(x')|e^{h_i(x) - h_i(x')} \frac{\sum_j e^{h_j(x')}}{\sum_j e^{h_j(x)}} - 1| \leq (\sum_i q^h_i(x')) 3 \hat{K} ||x' - x||_1$ for $ \hat{K} ||x' - x||_1 \leq 3/8$. Since $\sum_i q^h_i(x') = 1$, we have
$$
||q^h(x') - q^h(x)||_1 \leq 3 \hat{K} ||x' - x||_1 \mbox{ for }  ||x' - x||_1 \leq 3/8\hat{K}
$$
%

In other words $q^h$ is \emph{locally} $3\hat{K}$-Lipschitz for every $l_1$ norm ball of size $3/8\hat{K}$. The following allows us to prove global Lipschitzness.
\begin{lemma} \label{localtoglobal}
Any locally $L$-Lipschitz function $f$ for every $l_p$ ball of size $\delta_0$ on a compact convex set $X \subset \mathbb{R}^n$ is Lipschitz on the set $X$. The Lipschitz constant is also $L$.
\end{lemma}
\begin{proof}
Take any two points $x, y \in X$, the straight line joining $x,y$ lies in $X$ (as $X$ is convex). Also, a finite number of balls of size $\delta_0$ cover $X$ (due to compactness). Thus, there are finitely many points $x= z_1, \ldots, z_\mu = y$ on the line from $x, y$ such that $d_{l_p}(z_i, z_{i+1}) \leq \delta_0$. Further, since these points lie on a straight line we have 
$$
\begin{array}{l}
d_{l_p}(x, y) =  \sum_1^{\mu-1} d_{l_p}(z_i, z_{i+1})\end{array}
$$Then, let any metric $d$ be used to measure distance in the range space of $f$, thus, we get
$$
\begin{array}{l l}
d(f(x), f(y)) & \leq \sum_1^{\mu-1} d(f(z_i) , f(z_{i+1})) \\
&\leq  \sum_1^{\mu-1} L d_{l_p}(z_i, z_{i+1}) \\
& = L d_{l_p}(x, y)
\end{array}
$$
\end{proof}
Since in our case the defender mixed strategy space is compact and convex and $q^h(x)$ satisfies the above lemma with $L = 3\hat{K}$ and $\delta_0 = 3/8\hat{K}$, $q^h(x)$ is $3\hat{K}$-Lipschitz.

\end{proof}

%

\noindent\textbf{Proof of Theorem~\ref{paramutilitybound}}
\begin{proof}

Coupled with the guarantee that with prob. $\geq 1- \delta$, $Pr(||q^p(x) - q^{h}(x)||_1 \leq \sqrt{2} \Delta) \geq 1 - \Delta$, the assumptions guarantee that with prob. $\geq 1- \delta$ for the learned hypothesis $h$ there must exist a $x' \in B(x^*, \epsilon)$  such that 
$||q^p(x') - q^{h}(x')||_1 \leq \sqrt{2} \Delta$ and there must exist $x'' \in B(\tilde{x}, \epsilon)$  such that 
$||q^p(x'') - q^{h}(x'')||_1 \leq \sqrt{2} \Delta$. 

First, for notational ease let $\gamma$ denote $\sqrt{2} \Delta$.
The following are immediate using triangle inequality, with the results $||q^p(x') - q^{h}(x')||_1 \leq \gamma$ and $||q^p(x'') - q^{h}(x'')||_1 \leq \gamma$ and the Lipschitzness assumptions
$$
\begin{array}{c}
||q^p(x^*) -  q^{h}(x')||_1 \leq K \epsilon + \gamma \quad (\mbox{opt}x^*) \\
 ||q^p(\tilde{x}) -  q^{h}(x'')||_1 \leq 3\hat{K} \epsilon + \gamma \quad (\mbox{opt}\tilde{x})
 \end{array}
 $$
 We call $\tilde{x}^T U q^h(\tilde{x}) \geq x'^T U q^h (x') $ as equation opt$h$.
 Thus, we bound the utility loss as following
$$
\begin{array}{l}
 x^{*T} U q^p(x^*) - \tilde{x}^T U q^p(\tilde{x})  \\
 =  x^{*T} U q^p(x^*) - \tilde{x}^T U q^{h}(\tilde{x}) + \tilde{x}^T U q^{h}(\tilde{x}) -  \tilde{x}^T U p(y/\tilde{x}) \\
 \leq  x^{*T} U q^p(x^*) - x'^T U q^{h}(x') + \tilde{x}^T U q^{h}(\tilde{x}) -  \tilde{x}^T U p(y/\tilde{x}) \\
  \qquad \mbox { using }\mbox{opt}h\\
 = (x^* -x')^T U q^p(x^*) + x'^T U (q^p(x^*) - q^{h}(x')) + \\
 \quad \tilde{x}^T U q^{h}(\tilde{x}) -  \tilde{x}^T U q^p(\tilde{x}) \\
\leq \epsilon + (K\epsilon +  \gamma) + \tilde{x}^T U q^{h}(\tilde{x}) -  \tilde{x}^T U q^p(\tilde{x}) \\
\qquad \mbox{ using } x' \in  B(x^*, \epsilon), \mbox{opt}x^*\\
 =  ((K+1)\epsilon +  \gamma) + \tilde{x}^T  U (q^{h}(\tilde{x}) - q^{h}(x''))  + \\
 \quad \tilde{x}^T U (q^{h}(x'') -  q^p(\tilde{x}))\\
 \leq  (K+1)\epsilon +  \gamma + 6\hat{K} \epsilon + \gamma \\
 \qquad \mbox{ using } x'' \in  B(\tilde{x}, \epsilon) \mbox{ with Lipschitz }q^{h}, \mbox{opt}\tilde{x}
\end{array}
$$
\end{proof}

\section{Extension to Stackelberg Games} \label{app:extension}
Our technique extends to Stackelberg games by noting that the single resource case $K=1$ with $T-1$ targets gives
$\sum_{i=1}^{T-1} x_i \leq 1$. This directly maps to a probability distribution over $T$ actions.
The $x_i$'s with $x_{T} = 1 - \sum_{i=1}^{T-1} x_i$ is the probability of playing an action. With this set-up now the security game is a standard Stackelberg game, but where the leader has $T$ actions and follower has $T-1$ actions. 

Thus, in order to capture the general Stakelberg game, for the adversary, we assume $N$ actions for the adversary (instead of $T-1$ above). Then, similar to security games $q_1, \ldots, q_N$ denotes the adversary's  probability of playing an action. Thus, the function $h$ now outputs vectors of size $N-1$ (instead of $O(T)$), i.e., $A$ is a subset of $N-1$ dimensional Euclidean space. The model of security game in the PAC framework extends as is to this Stackelberg setup, just with $h(x)$ and $A$ being $N-1$ dimensional. The rest of the analysis proceeds exactly as for security games for both parametric and non-parametric case, by replacing the $T$ corresponding to the adversary's action space by $N$. Since, the proof technique is exactly same, we just state the final results. Thus, for a Stackelberg game with $T$ leader actions and $N$ follower actions, the bound for Theorem~\ref{PACbound} becomes
$$
\frac{576M^2}{\alpha^2} \log \frac{8\mathcal{C}(\alpha/96N, \mathcal{H}, d_{\bar{l_1}} )}{\delta}
$$
It can be seen from the proof for the parametric part that the sample complexity does not depend on the dimensionality of $X$, but only on the dimensionality of $A$.
Hence, the sample complexity results from generalized SUQR parametric case is 
$$
O\big(\frac{1}{\alpha^2} ( \log\frac{1}{\delta} +  N\log   \frac{N}{\alpha}    )\big)
$$
and for the non-parametric case, which depends on both dimensionality of $X$ and $T$, the sample complexity is
$$
O\big(\frac{1}{\alpha^2} ( \log\frac{1}{\delta} +  \frac{N^{T+1}}{\alpha^T}    )\big)
$$

\section{Analysis of Standard SUQR form} \label{app:stsuqr}
For SUQR the rewards and penalties are given and fixed.
Let the rewards be given and fixed $r = \langle r_1, \ldots, r_T \rangle$ (each $r_i \in [0, r_{\max}], r_{\max} > 0$), and the penalty values are $p = \langle p_1, \ldots, p_T \rangle$ (each $p_i \in [0, p_{\min}], p_{\min} < 0$). Thus, the output of $h$ is 
$$
\begin{array}{l}
h(x) = \\
\langle w_{1} x_{1T} + w_2 r_{1T} + w_3 p_{1T}, \ldots, \\w_{1}x_{T-1 T}  + w_2 r_{T-1T} + w_3 p_{T-1T} \rangle
\end{array}
$$
where $r_{iT} = r_i - r_T$ and same for $p_{iT}$. Note that in the above formulation all the component functions $h_i(x)$ have same weights. We can consider the function space $\mathcal{H}$ as the following direct-sum semi-free product $\mathcal{G} \oplus \mathcal{F} \oplus \mathcal{E} = \{\langle g_1+f_1+e_1, \ldots, g_{T-1}+f_{T-1}+e_{T-1} \rangle ~|~ \langle g_1, \ldots, g_{T-1} \rangle \in \mathcal{G}, \langle f_1, \ldots, f_{T-1} \rangle \in \mathcal{F}, \langle e_1, \ldots, e_{T-1} \rangle \in  \mathcal{E}\}$, where each of $\mathcal{G}, \mathcal{F}, \mathcal{E}$ is defined below. $\mathcal{G} = \{ \langle g_1, \ldots, g_{T-1} \rangle ~|~ \langle g_1, \ldots, g_{T-1} \rangle \in \times_i \mathcal{G}_i, \mbox{ all $g_i$ have same weight} \}$ where $\mathcal{G}_i$ has functions of the form $w x_{iT}$.  $\mathcal{F} = \{ \langle f_1, \ldots, f_{T-1} \rangle ~|~ \langle f_1, \ldots, f_{T-1} \rangle \in \times_i \mathcal{F}_i, \mbox{ all $f_i$ have same weight} \}$ where $\mathcal{F}_i$ has constant valued functions of the form $w r_{iT}$.  $\mathcal{E} = \{ \langle e_1, \ldots, e_{T-1} \rangle ~|~ \langle e_1, \ldots, e_{T-1} \rangle \in \times_i \mathcal{E}_i, \mbox{ all $e_i$ have same weight} \}$ where $\mathcal{E}_i$ has constant valued functions of the form $w p_{iT}$.

Consider an $\epsilon/3$-cover $U_e$ for $\mathcal{E}$, an $\epsilon/3$-cover $U_f$ for $\mathcal{F}$ and $\epsilon/3$-cover $U_g$ for $\mathcal{G}$. We claim that $U_e \times U_f \times  U_g$ is an $\epsilon$-cover for $\mathcal{E} \oplus \mathcal{F} \oplus \mathcal{G}$. Thus, the size of the $\epsilon$-cover for $\mathcal{E} \oplus \mathcal{F} \oplus \mathcal{G}$ is bounded by $|U_e| |U_f| |U_g|$.  Thus,
 $$
\mathcal{N}(\epsilon, \mathcal{H}, d_{\bar{l_1}}) \leq \mathcal{N}(\epsilon/3, \mathcal{G}, d_{\bar{l_1}}) \mathcal{N}(\epsilon/3, \mathcal{F}, d_{\bar{l_1}}) \mathcal{N}(\epsilon/3, \mathcal{E}, d_{\bar{l_1}})
$$
Taking sup over $P$ we get
 $$
\mathcal{C}(\epsilon, \mathcal{H}, d_{\bar{l_1}}) \leq \mathcal{C}(\epsilon/3, \mathcal{G}, d_{\bar{l_1}}) \mathcal{C}(\epsilon/3, \mathcal{F}, d_{\bar{l_1}}) \mathcal{C}(\epsilon/3, \mathcal{E}, d_{\bar{l_1}})
$$

Now, we show that $U_e \times U_f \times  U_g$ is an $\epsilon$-cover for $\mathcal{H} = \mathcal{E} \oplus \mathcal{F} \oplus \mathcal{G}$
Fix any $h \in \mathcal{H} = \mathcal{E} \oplus \mathcal{F} \oplus \mathcal{G}$. Then, $h = e + f + g$ for some $e \in \mathcal{E}, f \in \mathcal{F}, g \in \mathcal{G}$. Let $e' \in U_e$ be $\epsilon/3$ close to $e$, $f' \in U_f$ be $\epsilon/3$ close to $f$ and $g' \in U_g$ be $\epsilon/3$ close to $g$.

Then,
$$
\begin{array}{l}
\displaystyle d_{L^1(P, d_{\bar{l_1}})}(h, h')  \\  
\quad = \displaystyle\int_{X} \frac{1}{k} \sum_{i=1}^k d_{l_1}(h_i(x), h'_i(x)) \; dP(x) \quad\\
\quad \leq \displaystyle\int_{X} \frac{1}{k} \sum_{i=1}^k d_{l_1}(g_i(x), g'_i(x)) \\
\qquad \displaystyle + d_{l_1}(f_i(x), f'_i(x)) + d_{l_1}(e_i(x), e'_i(x)) \; dP(x) \\
\quad = d_{L^1(P, d_{\bar{l_1}})}(g, g') + d_{L^1(P, d_{\bar{l_1}})}(f, f') + d_{L^1(P, d_{\bar{l_1}})}(e, e')\\
\quad \leq \epsilon
\end{array}
 $$
 
 Similar to Lemma~\ref{Gbound}, it is possible to show that for any probability distribution $P$, for any function $g, g'$ $d_{\bar{l_1}}(g, g') \leq |w - w'|$ and $f, f'$ $d_{\bar{l_1}}(f, f') \leq |w - w'|r_{max}$ and $e, e'$ $d_{\bar{l_1}}(e, e') \leq |w - w'||p_{min}|$. Assume each of the functions have a range $[-M/6, M/6]$ (this does not affect the order in terms of $M$). Given, these ranges $w$ for $g$ can take values in $[-M/6, M/6]$, $w$ for $g$ can take values in $[-M/6r_{max}, M/6r_{max}]$ and $w$ for $g$ can take values in $[-M/6|p_{min}|, M/6|p_{min}|]$. To get a capacity of $\epsilon/3$ it is enough to divide the respective $w$ range into intervals of $2\epsilon/3$, and consider the boundaries. This yields an $\epsilon/3$-capacity of $M/2\epsilon$, $M/2\epsilon r_{max}$ and $M/2\epsilon |p_{min}|$ for $\mathcal{G}$, $\mathcal{F}$ and $\mathcal{E}$ respectively.
 
 Thus,
 $$
\mathcal{C}(\epsilon, \mathcal{H}, d_{\bar{l_1}}) \leq (M/2\epsilon)^3 \frac{1}{r_{max} |p_{min}|}
$$
Plugging this in sample complexity from Theorem~\ref{PACbound} we get the results that the sample complexity is
$$
O\big(\frac{1}{\alpha^2} ( \log\frac{1}{\delta} +  \log  \frac{T}{\alpha} )\big)
$$

\section{Alternate proof for Generalized SUQR Sample Complexity} \label{app:hausslerproof}

As discussed in the main paper we use the function space $\mathcal{H}'$ with each component function space $\mathcal{H}'_i$ given by $w_i x_{iT} + c_{iT}$. Then, we can directly use 
Equation~\ref{freebound}. We still need to bound $\mathcal{C}(\epsilon, \mathcal{H}'_i, d_{l_1})$. For this, we note the set of functions $w_i x_{iT} + c_{iT}$ has two free parameters $w_i$ and $c_i$, thus, this function space is a subset of the vector space of functions of dimension two (two values needs to represent each function). Using the pseudo-dimension technique~\cite{Haussler1992} we know that for psuedo-dimension $d$ of function space $\mathcal{H}_i$ we get
$$\mathcal{C}(\epsilon, \mathcal{H}'_i, d_{l_1}) \leq 2 (\frac{eM}{\epsilon} \log \frac{eM}{\epsilon})^d$$
Also, we know~\cite{Haussler1992} that pseudo-dimension is equal to the vector space dimension if the function class is a subset of a vector space. Therefore, for our case $d = 2$. Therefore, using Equation~\ref{freebound} we get
$$
\mathcal{C}(\epsilon, \mathcal{H}', d_{l_1}) \leq 2^T (\frac{eM}{\epsilon} \log \frac{eM}{\epsilon})^{2T}
$$
Plugging this result in Theorem~\ref{PACbound} we get the sample complexity of 
$$O\Big(\big(\frac{1}{\alpha^2}\big) \big ( \log (\frac{1}{\delta}) +  T\log (\frac{T}{\alpha} \log \frac{T}{\alpha})  \big)\Big)$$

\section{Experimental Results} \label{app:experiments}

\begin{figure*}[!hbt]
    \centering
	\subfigure[AMT Parametric Results Payoff 2]{
    \label{fig:AMT_param_payoff2}
    \includegraphics[width = 0.22\textwidth, height=2cm]{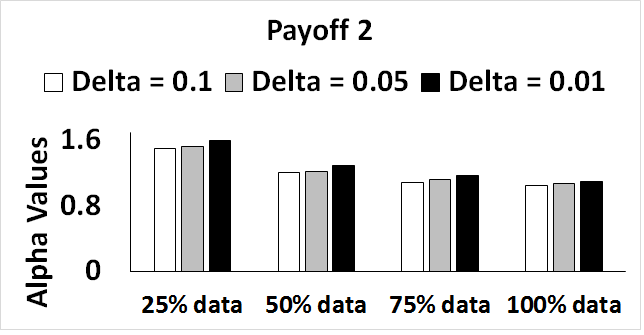}
    }	
	\subfigure[AMT Nonparametric Results Payoff 2]{
    \label{fig:AMT_nonparam_payoff2}
    \includegraphics[width = 0.22\textwidth, height=2cm]{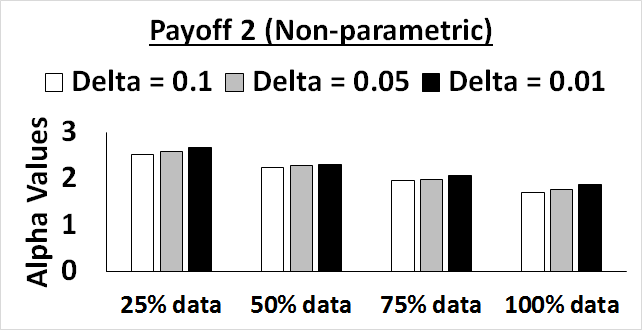}
    }	
    \subfigure[Simulated Data Payoff 1 - Parametric results]{
    \label{fig:SimPayoff1_param}
    \includegraphics[width = 0.22\textwidth, height=2cm]{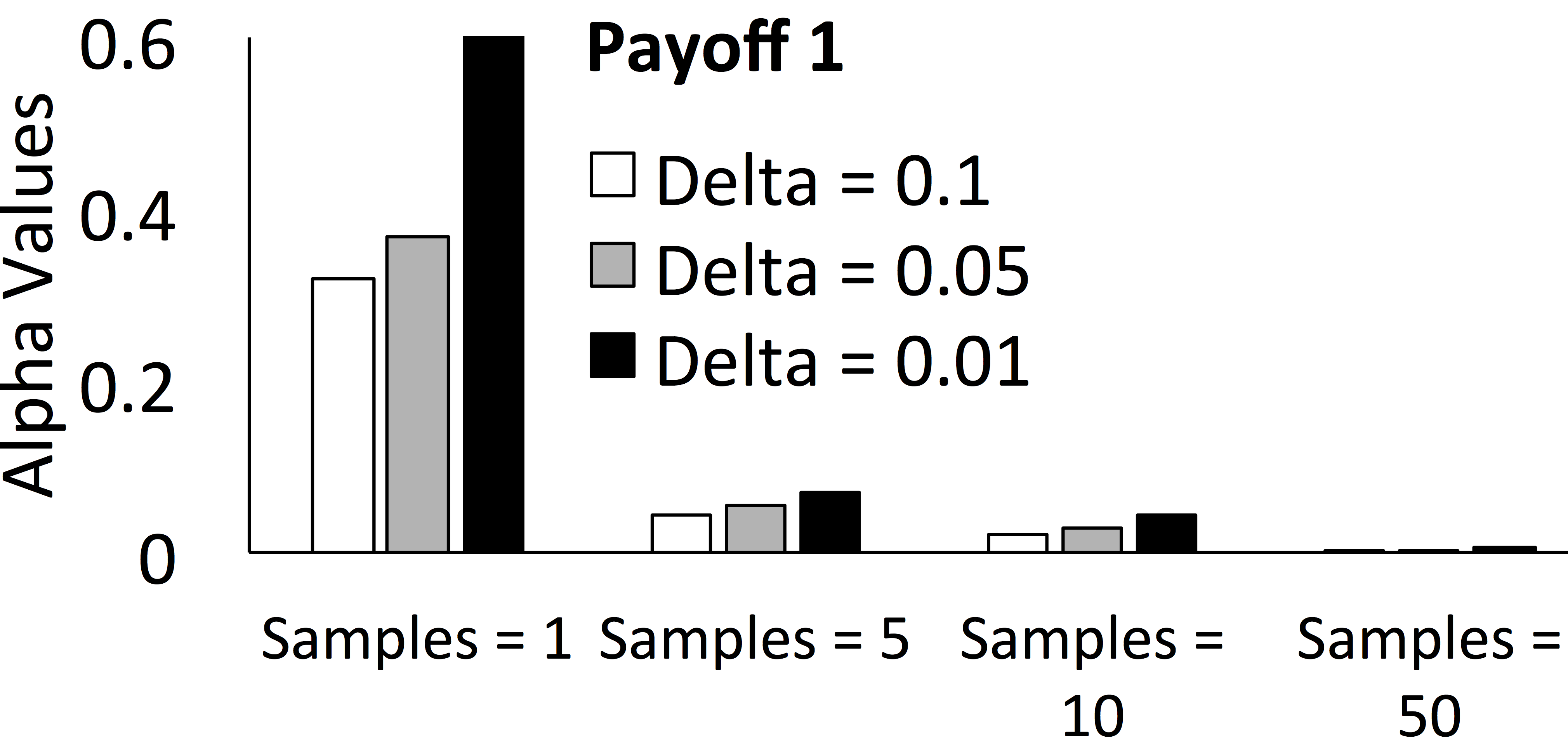}
    }
   \subfigure[Simulated Data Payoff 2 - Parametric results]{
    \label{fig:SimPayoff2_param}
    \includegraphics[width = 0.22\textwidth, height=2cm]{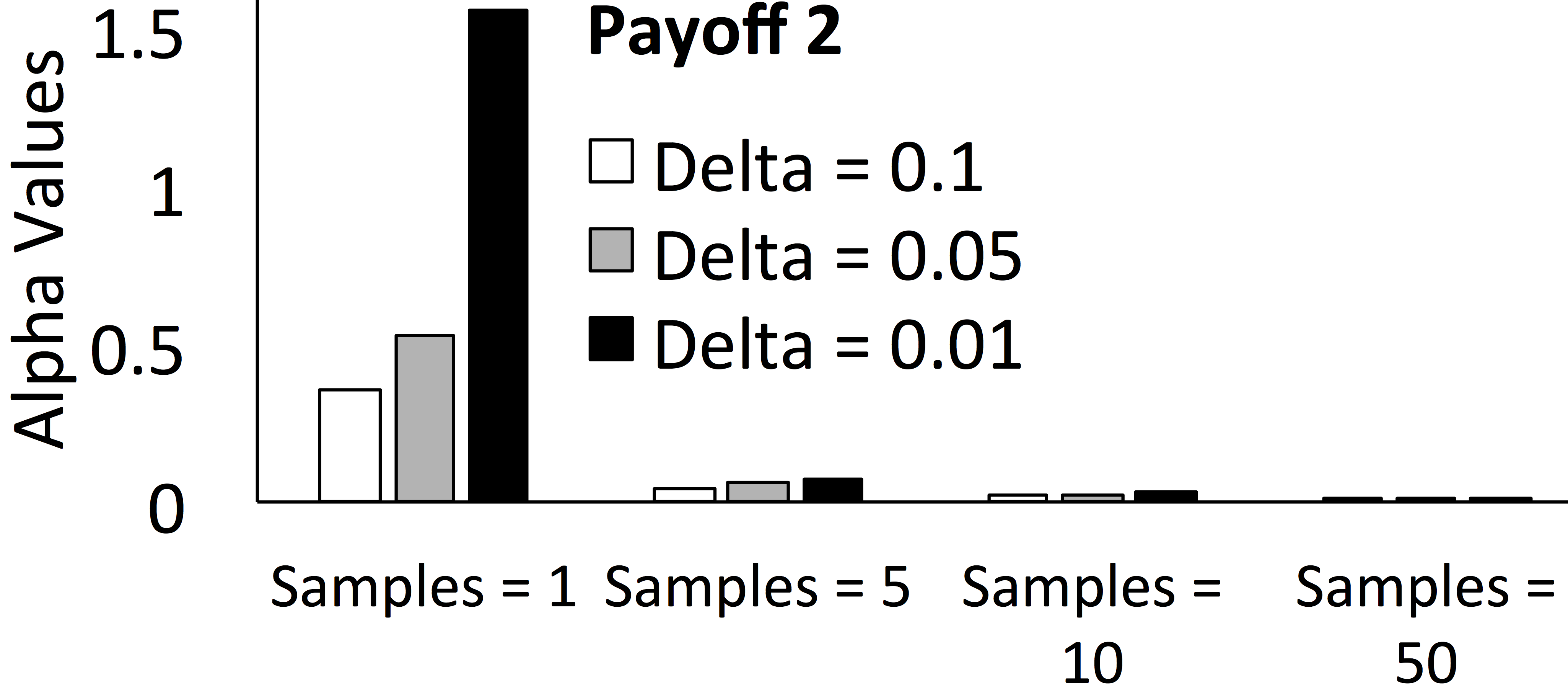}
    }
	\subfigure[Simulated Data Payoff 1 - Nonparametric results]{
    \label{fig:SimPayoff1_nonparam}
    \includegraphics[width = 0.22\textwidth, height=2cm]{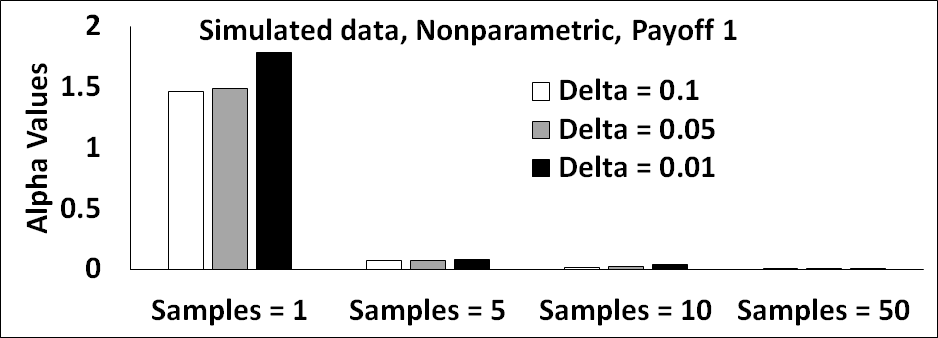}
    }
   \subfigure[Simulated Data Payoff 2 - Nonparametric results]{
    \label{fig:SimPayoff2_nonparam}
    \includegraphics[width = 0.22\textwidth, height=2cm]{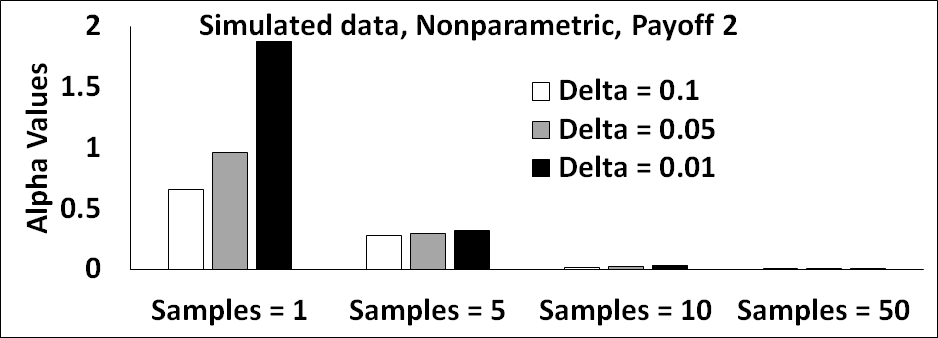}
    }
	\subfigure[Parametric vs Non-parametric results on Simulated (for various sample sizes) data from payoff 1 when the true adversary model is different from the parametric learned function]{
    \label{fig:DiffModel_Payoff1}
    \includegraphics[width = 0.48\textwidth, height=3cm]{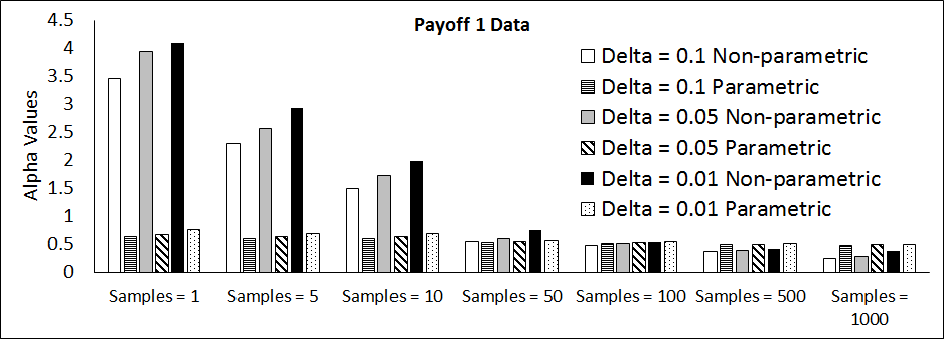}
    }
    \subfigure[Parametric vs Non-parametric results on Simulated (for various sample sizes) data from payoff 2 when the true adversary model is different from the parametric learned function]{
    \label{fig:DiffModel_Payoff2}
    \includegraphics[width = 0.48\textwidth, height=3cm]{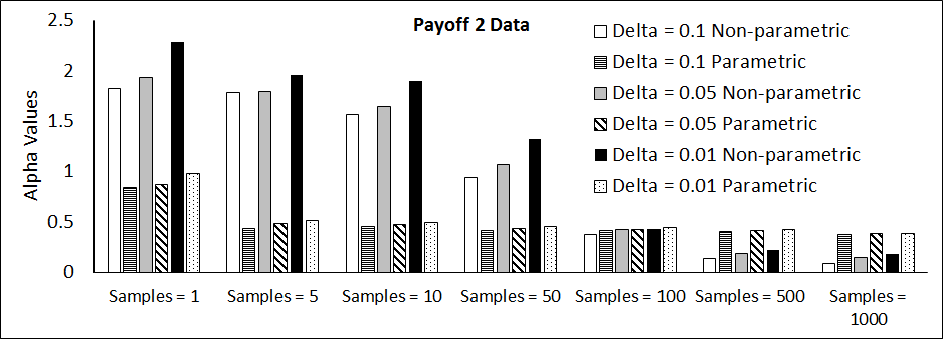}
    }
    \caption{Results on Uganda, AMT and simulated datasets for the parametric and non-parametric cases respectively.}
\end{figure*}

Here we provide additional experimental results on the Uganda, AMT and simulated datasets. 
The AMT dataset consisted of 32 unique mixed strategies, 16 of which were deployed for one payoff structure and the remaining 16 for another. In the main paper, we provided results on AMT data for payoff structure 1. Here, in Figs. \ref{fig:AMT_param_payoff2} and \ref{fig:AMT_nonparam_payoff2}, we show results on the AMT data for both the parametric (SUQR) and NPL learning settings on payoff structure 2. 

For running experiments on simulated data, we used the same mixed strategies and features as for the AMT data, but simulated attacks, first using the actual SUQR model and then using a modified form of the SUQR model. Figs. \ref{fig:SimPayoff1_param} and \ref{fig:SimPayoff2_param} show results on simulated data on payoff structures 1 and 2 for the parametric cases, when the data is generated by an adversary with an SUQR model with true weight vector reported in Nguyen et. al~\shortcite{Nguyen13analyzingthe} ($(w_1, w_2, w_3)=(-9.85, 0.37, 0.15)$ ($c_i = w_2 R_i + w_3 P_i$)). Similar results for the NPL model are shown in Figs. \ref{fig:SimPayoff1_nonparam} and \ref{fig:SimPayoff2_nonparam} respectively. We can see that the NPL approach performs poorly with only one or five samples as expectied but improves significantly as more samples are added. To further show its potential, we modified the true adversary model of generating attacks from SUQR to the following: $q_i \propto e^{w_1 x_i^2 + c_i}$, i.e., instead of $x_i$, the adversary reasons based on $x_i^2$. We considered the same true weight vector to simulate attacks. Then, we observe in Figs. \ref{fig:DiffModel_Payoff1} (for payoff structure 1) and \ref{fig:DiffModel_Payoff2} (for payoff structure 2 data), that $\alpha$ approaches a value closer to zero for 500 or more sample. Also, the NPL model performs better than the parametric model with 500 or more samples. This shows that the NPL approach is more accurate when the true adversary does not satisfy the simple parametric logistic form, indicating that when we don't know the true function of the adversary's decision making process, adopting a non-parametric method to learn the adversary's behavior is more effective.

\end{document}